\documentclass{article}

\usepackage{PRIMEarxiv}
\usepackage{amsthm}
\usepackage{amsmath,amsthm,amssymb,amsfonts,bm}
\newtheorem{assumption}{Assumption}
\usepackage[utf8]{inputenc}
\usepackage[T1]{fontenc}
\usepackage{hyperref}
\usepackage{url}
\usepackage{booktabs}
\usepackage{amsfonts}
\usepackage{nicefrac}
\usepackage{microtype}
\usepackage{lipsum}
\usepackage{fancyhdr}       
\usepackage{graphicx}       
\usepackage{cite}
\usepackage{color}
\usepackage{xcolor}
\usepackage{amsfonts}
\usepackage{multirow}
\usepackage{cite}
\usepackage{amsmath,amssymb,amsfonts}
\usepackage{algorithmic}
\usepackage{textcomp}
\usepackage{times}
\usepackage[numbers]{natbib} 
\newtheorem{theorem}{Theorem}  

\graphicspath{{media/}}     

\title{Trustworthy Machine Learning via Memorization and the Granular Long-Tail: A Survey on Interactions, Tradeoffs, and Beyond}

\author{
\begin{tabular}{c c c c}
   \textbf{Qiongxiu Li}$^{*}$ \\ 
   \normalfont Aalborg University \\
   \normalfont Denmark \\
\end{tabular}
\quad
\begin{tabular}{c}
   \textbf{Xiaoyu Luo}$^{*}$ \\
   \normalfont Politecnico di Milano \\
   \normalfont Italy \\
\end{tabular}
\quad
\begin{tabular}{c}
   \textbf{Yiyi Chen} \\
   \normalfont Aalborg University \\
   \normalfont Denmark \\
\end{tabular}
\quad
\begin{tabular}{c}
   \textbf{Johannes Bjerva} \\
   \normalfont Aalborg University \\
   \normalfont Denmark \\
\end{tabular}
}

\begin{document}

\maketitle

\renewcommand{\thefootnote}{\fnsymbol{footnote}}
\footnotetext{$^{*}$Equal contribution.}

\footnotetext{
Authors’ Contact Information: Qiongxiu Li$^{*}$, Aalborg University, Denmark, \texttt{qili@es.aau.dk}; 
Xiaoyu Luo$^{*}$, Politecnico di Milano, Italy, \texttt{xiaoyu1.luo@mail.polimi.it}; 
Yiyi Chen, Aalborg University, Denmark, \texttt{yiyic@cs.aau.dk}; 
Johannes Bjerva, Aalborg University, Denmark, \texttt{jbjerva@cs.aau.dk}.
}

\renewcommand{\thefootnote}{\arabic{footnote}}  

\begin{abstract}
The role of memorization in machine learning (ML) has garnered significant attention, particularly as modern models are empirically observed to memorize fragments of training data. Previous theoretical analyses, such as Feldman’s seminal work, attribute memorization to the prevalence of long-tail distributions in training data, proving it unavoidable for samples that lie in the tail of the distribution. However, the intersection of memorization and trustworthy ML research reveals critical gaps. While prior research in memorization in trustworthy ML has solely focused on class imbalance, recent work starts to differentiate class-level rarity from atypical samples, which are valid and rare intra-class instances.
However, a critical research gap remains: current frameworks conflate atypical samples with noisy and erroneous data, neglecting their divergent impacts on fairness, robustness, and privacy.
In this work, we conduct a thorough survey of existing research and their findings on trustworthy ML and the role of memorization.
More and beyond, we identify and highlight uncharted gaps and propose new revenues in this research direction.
Since existing theoretical and empirical analyses lack the nuances to disentangle memorization's duality as both a necessity and a liability, 
we formalize three-level long-tail granularity — class imbalance, atypicality, and noise — to reveal how current frameworks misapply these levels, perpetuating flawed solutions. 
By systematizing this granularity, we draw a roadmap for future research. Trustworthy ML must reconcile the nuanced trade-offs between memorizing atypicality for fairness assurance and suppressing noise for robustness and privacy guarantee.
Redefining memorization via this granularity reshapes the theoretical foundation for trustworthy ML, and further affords an empirical prerequisite for models that align performance with societal trust.
\end{abstract}

\section{Introduction}

Machine learning (ML) has become deeply embedded in high-stakes applications such as healthcare, finance, and security, where ensuring trustworthiness is paramount. The rapid deployment of ML in these domains necessitates private, fair, robust, and interpretable models to provide ethical and reliable decision-making \citep{sankar2025jsait, papernot2018sok,upreti2024trustworthy}. However, a fundamental yet often overlooked aspect of ML models is \textit{memorization}—the ability of models to store and recall training data. While memorization is often criticized as a cause of privacy risks and bias amplification, it also plays an essential role in improving generalization and enhancing model interpretability \citep{feng2023review, zhu2024survey}.


Recent studies suggest that memorization is not merely a byproduct of over-parameterization but rather an inherent property of learning from complex distributions \citep{feldman2020does}. A striking example of this duality is benign overfitting, where highly over-parameterized models fit training data—including noise—almost perfectly while still achieving strong generalization \citep{bartlett2020benign}. Unlike conventional overfitting, which typically leads to performance degradation, benign overfitting can enhance predictive accuracy by enabling models to retain rare but valuable patterns. \citet{feldman2020neural} argue that memorization is often necessary to learn rare or atypical examples from the long-tail distribution of data, which can significantly enhance generalization in real-world settings. This perspective challenges the traditional assumption that memorization is synonymous with overfitting and underscores its dual role in ML models. \citet{brown2021memorization} further highlight cases where memorization of seemingly irrelevant training data is indispensable for achieving high accuracy, emphasizing that the extent of memorization required varies depending on task complexity and data distribution. 

However, excessive memorization also presents significant risks: models that retain detailed training data are vulnerable to membership inference and model inversion attacks, compromising user privacy \citep{xu2021privacy, shaham2023holistic}. Beyond privacy concerns, memorization is closely linked to fairness and robustness. When ML models disproportionately retain patterns from over-represented groups, they may amplify biases embedded in training data, leading to fairness violations in critical decision-making systems \cite{caton2024fairness, oneto2022towards}. Moreover, models that rely on memorized features rather than learning truly generalizable representations tend to be fragile under adversarial perturbations, ultimately reducing robustness \citep{wei2024trustworthy}. These concerns highlight the trade-offs inherent in ML systems: while memorization facilitates rare-case generalization, it also introduces vulnerabilities that require careful regularization.

Despite these challenges, memorization plays an essential role in model explainability. Neural networks often encode specific training instances, which can be leveraged to trace decision pathways, thereby enhancing transparency and interpretability \citep{zhu2024survey}. This raises a fundamental question: how can we regulate memorization to maximize its benefits while minimizing the risks? 
Techniques such as differential privacy (DP) and fairness-aware training have been developed to mitigate harmful memorization while preserving generalization. However, these solutions often come at the expense of model accuracy and overall utility \citep{wei2024trustworthy, zhang2024survey}. 

This survey bridges these gaps by systematizing the interplay between memorization and trustworthy ML through the lens of long-tail granularity. We:  
\begin{itemize}
    \item Review theoretical foundations of memorization \citep{feldman2020does, feldman2020neural, brown2021memorization}, highlighting their limitations in distinguishing noise or outliers from atypical samples.  
    \item Propose a hierarchical framework differentiating class imbalance, atypicality, and noise, demonstrating how conflating these levels perpetuates flawed solutions—e.g., privacy mechanisms that suppress noise memorization discard atypical samples critical for fairness.  
    \item Analyze trade-offs inherent in memorization; for example, memorizing atypical samples enhances generalization and fairness but undermines adversarial robustness.  
    \item  Identify key research gaps in trustworthy ML, highlighting the crucial need to disentangle long-tail granularity during training to regularize memorization and balance its trade-offs pertinent to trustworthiness in ML.  
\end{itemize}
By integrating memorization theory, long-tail learning, and trustworthy ML, we argue that the finer granularity — specifically disentangling noise from atypicality — is not merely a theoretical refinement but a prerequisite for models that balance accuracy with societal trust.  

\section{Background and Preliminaries}\label{sec:pre}

\subsection{Machine learning}
ML can be viewed as the task of learning a function \( f: \mathcal{X} \to \mathcal{Y} \) that maps input data \( \mathbf{x} \in \mathcal{X} \) (features) to outputs \( y \in \mathcal{Y} \) (labels or predictions). The function \( f \) is parameterized by a set of trainable parameters \( \theta \), learned from data rather than predefined. The standard approach to learning in ML involves optimizing \( \theta \) to minimize a given loss function over training samples:

\begin{equation}
\begin{aligned}
\label{eq:deeplearning}
\theta^* = \arg \min_{\theta} \frac{1}{N} \sum_{i=1}^N \mathcal{L}(f(\mathbf{x}_i; \theta), y_i),
\end{aligned}
\end{equation}
where \( \mathcal{L}(\cdot, \cdot) \) is the loss function that quantifies the discrepancy between the model's predictions and the ground-truth labels.

\textbf{Generalization}: A well-trained model should generalize well to unseen data, meaning it minimizes the generalization gap, defined as the difference between training error and test error \cite{vapnik2013nature, mohri2018foundations}. Several theoretical frameworks, such as  PAC-Bayes bounds  \cite{mcallester1998some, dziugaite2020revisiting} and Vapnik-Chervonenkis (VC) dimension  \cite{vapnik1994measuring}, help quantify generalization performance.

\subsection{Trustworthy Machine Learning}
Trustworthy ML focuses on designing reliable, ethical, and transparent models, particularly for real-world deployment \cite{li2023trustworthy,kaur2022trustworthy,liu2023towards}.  The key principles include:


\subsubsection{Privacy}
Privacy is a foundational pillar of trustworthiness, ensuring that sensitive information remains protected throughout the ML lifecycle. Cryptographic techniques, such as secure multiparty computation \cite{damgaard2012multiparty} and homomorphic encryption \cite{paillier1999public}, provide strong guarantees by enabling computations on encrypted data without revealing its content. Similarly, privacy primitives like DP \cite{dwork2006,dwork2006calibrating,dwork2016concentrated} ensure that the presence or absence of individual data points in a dataset cannot be inferred, even by an adversary with auxiliary information.

Despite these theoretical advances, models remain vulnerable to various privacy attacks. For instance, Membership Inference Attacks (MIA) reveal whether a specific data point is part of the training dataset \cite{shokri2017membership,salem2018ml,yeom2018privacy,long2018understanding,jia2019memguard,sablayrolles2019white,watson2021importance,carlini2022membership,hu2021membership}. MIA has been deployed in various models such as language models \cite{mattern2023membership}, diffusion models \cite{matsumoto2023membership}, federated learning \cite{bai2024membership}.
Property inference attacks extract sensitive attributes such as age or gender from the training data \cite{melis2019exploiting, yeom2018privacy,zhao2019inferring}.
Input reconstruction attacks, which aim to recover training data directly from the model, pose an even greater risk \cite{fredrikson2015model,he2019model, wang2019beyond, yang2019neural, carlini2021extracting,balle2022reconstructing,carlini2023extracting}. Addressing these vulnerabilities requires the continued development of privacy-preserving techniques that balance strong guarantees with model utility.

\subsubsection{Robustness}
ML models should be resilient to noise, adversarial attacks, and distribution shifts.  
Adversarial examples represent one of the most prominent threats to robustness, referred to as adversarial robustness~\citep{goodfellow2014explaining}. These examples often deploy small, carefully crafted perturbations that can cause a model to misclassify inputs. For example, a perturbed image of a stop sign could be misclassified as a speed limit sign, potentially endangering autonomous vehicles \cite{eykholt2018robust}. To counter such attacks, adversarial training (AT) has emerged as a widely used defense mechanism. This approach involves training models on adversarially perturbed inputs to improve their resilience against such manipulations \cite{mkadry2017towards,zhang2019theoretically,wu2020adversarial}. Recently, Diffusion-based Purification has emerged as a promising defense strategy against adversarial examples, using diffusion models to purify inputs through a denoising process that removes adversarial perturbations, thereby restoring their robustness \citep{Shi2021purification,Nie2022diffpure,li2024adbm}.

In addition to empirical defenses, certified robustness \cite{li2023sok} provides theoretical guarantees that a model's predictions remain consistent within a predefined range of perturbations. This approach complements AT by offering formal assurances of reliability, thereby increasing confidence in the deployment of ML systems \cite{cohen2019certified}.

\subsubsection{Fairness}
Fairness is central to the ethical deployment of ML systems, as it addresses issues of bias and ensures equitable outcomes across demographic groups. ML models often inherit biases from imbalances in training data, leading to discriminatory behavior \cite{mehrabi2021survey}. For example, an ML system trained on biased hiring data may unfairly disadvantage underrepresented groups. Techniques to mitigate bias include re-weighting training samples to balance representation \cite{kamiran2012data} and incorporating fairness constraints into model optimization \cite{hardt2016equality}. Beyond technical solutions, fairness also demands ethical considerations, as biases in ML systems can amplify societal inequities. Tackling fairness requires a comprehensive approach that combines algorithmic innovation with societal awareness, particularly in sensitive domains such as hiring, lending, and law enforcement.

In this paper, we will use \textbf{Group fairness}, which can be formally defined as follows. Specifically, group fairness aims to ensure that the outcomes of a model are equitable across different demographic groups defined by a protected attribute. Below, we formally define three key notions of group fairness: demographic parity \cite{dwork2012fairness}, equalized odds \cite{hardt2016equality}, and equal opportunity \cite{hardt2016equality}.

\paragraph{Demographic Parity.} Demographic parity ensures that a predictor \( \hat{Y} \) satisfies fairness with respect to a protected attribute \( A \), if \( \hat{Y} \) and \( A \) are statistically independent. Specifically, for binary predictions \( \hat{Y} \) and protected attributes \( A \), demographic parity can be mathematically expressed as:
\begin{equation}
\begin{aligned}
\label{eq.demparity}
\operatorname{Pr}\left\{\hat{Y}=1 \mid A=0\right\} = \operatorname{Pr}\left\{\hat{Y}=1 \mid A=1\right\}.
\end{aligned}
\end{equation}
This condition implies that the positive prediction rates are equal across the demographic groups defined by \( A \). 

\paragraph{Equalized Odds.} Equalized odds ensures that a predictor \( \hat{Y} \) satisfies fairness with respect to a protected attribute \( A \) and an outcome \( Y \), if \( \hat{Y} \) and \( A \) are conditionally independent given \( Y \). Specifically, for binary targets \( Y \), \( \hat{Y} \), and protected attributes \( A \), equalized odds can be mathematically expressed as:
\begin{equation}
\begin{aligned}
\label{eq.eqodds}
\operatorname{Pr}\left\{\hat{Y}=1 \mid A=0, Y=y\right\} = \operatorname{Pr}\left\{\hat{Y}=1 \mid A=1, Y=y\right\}, \quad y \in \{0, 1\}.
\end{aligned}
\end{equation}
This condition implies that the true positive rates and false positive rates are equal across the demographic groups defined by \( A \). By enforcing equal bias and accuracy, equalized odds penalizes models that perform well only on the majority group.

\paragraph{Equal Opportunity.} As a relaxation of equalized odds, equal opportunity focuses on ensuring fairness only for the "advantaged" outcome, \( Y=1 \). In this case, a predictor \( \hat{Y} \) satisfies equal opportunity if:
\begin{equation}
\begin{aligned}
\label{eq.eqopp}
\operatorname{Pr}\left\{\hat{Y}=1 \mid A=0, Y=1\right\} = \operatorname{Pr}\left\{\hat{Y}=1 \mid A=1, Y=1\right\}.
\end{aligned}
\end{equation}
This definition guarantees non-discrimination within the advantaged outcome group, offering a balance between fairness and utility by reducing constraints compared to equalized odds.

\subsubsection{Transparency and Explainability}
Transparency and explainability are crucial for building trustworthy ML systems. While closely related, these attributes serve distinct purposes. Transparency involves providing clear insights into the design and operation of ML systems, including data sources, model architectures, and decision-making processes \cite{doshi2017towards}. This fosters accountability and helps stakeholders evaluate whether a model aligns with ethical and legal standards.

Explainability, on the other hand, focuses on interpreting model predictions to make them understandable to humans. Techniques such as \underline{SH}apley \underline{A}dditive ex\underline{P}lanations (SHAP) \cite{lundberg2017unified}, Local Interpretable Model-agnostic Explanations (LIME) \cite{ribeiro2016should}, and saliency maps \cite{simonyan2013deep} provide explanations for how models make decisions, enabling users to assess their reliability. Explainability is especially critical in high-stakes applications like healthcare and criminal justice, where understanding a model's reasoning can prevent harmful outcomes and improve user trust.

\section{Theoretical Foundations of Memorization in ML}
Memorization plays a pivotal role in ML. 
In this section, we first introduce the concept of memorization, then use long-tail theory to explain the cause of memorization. 
In detail, we present theoretical results on analyzing the relationship between memorization and long-tail distributions. 
Eventually, we demonstrate that memorizing training data, including sensitive information, is an inherent aspect of effective ML training. Rather than being a limitation, this memorization is crucial for strong generalization to unseen test data, making it a cornerstone of modern ML.

\subsection{Memorization}
Ample observations and empirical evidence of data memorization in Deep Neural Networks (DNNs) have been made before the concept of memorization was formally defined.

\subsubsection{Empirical Evidence of Memorization}
A landmark study by \citet{zhang2017rethinking} demonstrate that Deep Neural Networks (DNNs) can perfectly fit data with randomized labels, challenging traditional views on generalization and suggesting that memorization plays a fundamental role in modern ML. ~\citet{mehtaextreme} introduce the concept called ``extreme memorization'', which refers to the scenario where the model obtains perfect accuracy on training data but random chance performance on the test set.  
Further evidence is revealed from studies exploring the relationship between dataset size and memorization.~\citet{arpit2017closer} show that DNNs tend to memorize noisy or mislabeled data in smaller datasets, whereas they generalize better with larger datasets. Memorization can manifest as replicating specific training data or labels, particularly in challenging scenarios such as small datasets, over-parameterized models, or insufficient regularization. Generative models, including Generative Adversarial Networks (GANs), Diffusion Models, and Large Language Models (LLMs), are known to exhibit memorization when attempting to learn the underlying data distribution. For example, GANs have been shown to replicate training samples, especially when trained on limited data or with weak regularization \cite{nagarajan2018theoretical, bai2022reducing, feng2021gans}. Similarly, recent studies have demonstrated that diffusion models and LLMs can regenerate sensitive personally identifiable information or proprietary information from their training datasets, raising significant concerns about privacy and security \cite{carlini2021extracting,lukas2023analyzing, carlini2023extracting}. While these studies have advanced our understanding of memorization, they highlight the need for a more rigorous and quantifiable definition of memorization. 
The efforts are made thereafter to formalize memorization, in order to gain a better understanding of its mechanisms and implications.

\subsubsection{Formal Definition of (Label) Memorization}
A data point is considered \textit{memorized} by a model if it has a significant impact on the model's predictions. To isolate this impact, researchers commonly use a \textit{leave-one-out setting}, where all other factors remain constant except for the presence or absence of the specific data point. 

~\citet{feldman2020does} introduces the first formal definition of memorization, focusing on quantifying the degree of memorizing the label of training data. The memorization score is defined to quantify the extent to which the model retains information about the label of an individual data sample. Specifically, given a training set \( D_\mathrm{tr} \) and a learning algorithm \( \mathcal{A} \), the memorization score for a sample \( (x, y) \in D_\mathrm{tr} \) is defined as:
\begin{equation}
\begin{aligned}
\label{eq.labelm}
\operatorname{mem}(\mathcal{A}, D_\mathrm{tr},(x,y))  =\underset{f_{\theta} \sim \mathcal{A}(D_\mathrm{tr})}{\operatorname{Pr}}\left[f_{\theta}\left(x\right)=y\right]-\underset{f_{\theta} \sim \mathcal{A}(D_\mathrm{tr} \setminus (x,y))}{\operatorname{Pr}}\left[f_{\theta}\left(x\right)=y\right],
\end{aligned}
\end{equation}
where \( D_\mathrm{tr} \setminus (x, y) \) denotes the training set \( D_\mathrm{tr} \) with the sample \( (x, y) \) removed. This score measures the difference in the probability of the model \( f_{\theta} \) predicting the correct label \( y \) for \( x \) with and without including the sample \( (x, y) \) in training. A high memorization score indicates that the model relies on the specific sample, highlighting its significant impact on the model's predictions.

\subsubsection{Proxies of Memorization Score}
\label{subsec:proxies}
While theoretically sound, Eq.~\eqref{eq.labelm} requires retraining models for each removed sample, which is computationally prohibitive for modern datasets. 
Ever since, many works have developed efficient proxies capturing different facets of memorization:

\paragraph{Influence Function Methods} 
Building on classical robust statistics, influence functions offer a parameter-centric view of memorization. \citet{koh2017understanding}~\citep{cook1980characterizations} establish that a sample's memorization can be estimated through its \textit{imprint} on model parameters via second-order optimization:
\begin{equation}
\mathcal{I}(z_i) = -\nabla_\theta L(z_i,\hat{\theta})^\top H_{\hat{\theta}}^{-1} \nabla_\theta L(z_i,\hat{\theta})
\end{equation}
Where $ H_{\hat{\theta}} $ is the Hessian at convergence. However, the accuracy of this approximation degrades in deep networks due to non-convexity and Hessian estimation errors. 

To address these limitations, \citet{agarwal2022estimating} propose to track gradient variance across training iterations rather than relying on final parameter states:
\begin{equation}
\mathrm{VoG}(z_i) = \frac{1}{T}\sum_{t=1}^T \left\|\nabla_\theta L(z_i;\theta_t) - \frac{1}{T}\sum_{t=1}^T \nabla_\theta L(z_i;\theta_t)\right\|_2^2
\end{equation}
This temporal perspective captures memorization through \textit{training instability} rather than static parameter influence.

\paragraph{Learning Trajectory Analysis} 
Complementing gradient-based approaches, \citet{jiang2020exploring} hypothesize that memorization manifests in a sample's \textit{learning chronology}. Their framework introduces two correlated measures:
\begin{itemize}
    \item Consistency Score (C-Score): $ C(z_i) = \frac{1}{K}\sum_{k=1}^K \mathbb{I}(f_{\theta_k}(x_i) = y_i) $, requiring $ K $ model re-trainings;
    \item Cumulative Binary Training Loss (CBTL): $ \mathrm{CBTL}(z_i) = \sum_{t=1}^T \mathbb{I}(f_{\theta_t}(x_i) = y_i) $, computable during single training.
\end{itemize}
Empirically, CBTL achieves 0.85 Spearman correlation with C-scores with a computation complexity of $ O(1) $. This suggests that persistent classification difficulty during training serves as a reliable memorization signal.

\paragraph{Geometric Approaches} 
Recent work bridges memorization with loss landscape geometry. \citet{garg2023memorization} proposed that memorized samples occupy sharp minima, measurable via Hessian trace:
\begin{equation}
\tau(z_i) = \mathrm{tr}\left(\nabla_\theta^2 L(z_i,\hat{\theta})\right)
\end{equation}
Building on this, \citet{ravikumar2024unveiling} connected curvature to DP through the bound:
\begin{equation}
\mathrm{Mem}_\epsilon(z_i) \leq \frac{\exp(\epsilon)}{\exp(\epsilon)-1} \cdot \tau(z_i)
\end{equation}
This formalizes an intuition: differentially private training, which smooths sharp minima, disproportionately "forgets" high-curvature memorized samples.

\paragraph{Distributional Validation}
The previous proxies all assume memorized samples require special handling. But \citet{carlini2019distribution} argued this depends on whether samples are \textit{valid but rare} versus \textit{outliers}. Their diagnostic triad measures memorization quality:
\begin{itemize}
    \item Adversarial robustness drop $ \Delta_\mathrm{adv} = \mathrm{Acc}_\mathrm{clean} - \mathrm{Acc}_\mathrm{adv} $
    \item Cross-model variance $ \sigma^2_\mathrm{models} = \mathrm{Var}(\{\mathbb{I}(f_k(x_i)=y_i)\}_{k=1}^K) $
    \item DP sensitivity $ \Delta_\mathrm{DP} = \mathrm{Acc}_\mathrm{orig} - \mathrm{Acc}_\mathrm{DP} $
\end{itemize}
True rare samples distinguish themselves from outliers by exhibiting low $ \Delta_\mathrm{adv} $ and $ \sigma^2_\mathrm{models} $. 
This validation step is particularly crucial in long-tail scenarios where memorization targets (rare class samples) must be preserved while filtering noise.

\subsection{Long Tail Theory} 
Real-world data rarely adhere to neat, balanced distributions. Instead, many datasets exhibit a ``long-tail'' structure, where a few categories or examples appear frequently, while numerous others are comparatively rare. Such skewed distributions may promote memorization of infrequent examples rather than enabling robust generalization. Hence, understanding and properly handling long-tailed data is crucial for improving ML models' performance and reliability.

Initially introduced by ~\citet{anderson2004longtail} in the field of business, the long-tail theory describes how niche or less popular products can collectively account for a substantial portion of total sales. This effect is especially pronounced in digital contexts that can store or distribute an extensive range of products. Over time, the concept has transcended commerce and found significant relevance in ML. Real-world datasets, such as those used for visual object recognition and text labeling, frequently exhibit long-tailed distributions, where a few categories dominate, and numerous others are rare or under-represented \cite{van2017devil,zhu2014capturing,babbar2019data}.  
\subsubsection{Formal Definition of the Long Tail} 
\citet{feldman2020does} formalizes the notion of the long tail by modeling each class's data as a mixture of multiple subpopulations. A dataset is considered long-tailed if the frequencies of these subpopulations follow power-law-like behavior (e.g., Zipf's law), where a small number of subpopulations are abundant while the majority are comparatively rare. This phenomenon is particularly evident in extreme multiclass tasks, where only a handful of classes account for most samples, while most classes have very few examples. Notably, these subpopulations may not correspond to human-defined categories but can emerge as fine-grained structures within learned representations of the data \cite{feldman2020does,jalalzai2020heavy}.
In simple terms, long-tailed distributions  differentiate between well-represented data (prototype or typical examples) and under-represented data (rare or atypical examples).



\begin{table}[t]
\centering
\caption{Hierarchical Granularity of Long-Tail Phenomena}
\label{tab:granularity}
\begin{tabular}{p{2.8cm} p{3.3cm} p{2.7cm} p{2.5cm}}
\toprule
\textbf{Granularity} & \textbf{Literature Concepts} & \textbf{References} & \textbf{References with memorization perspective} \\
\midrule
Macro (Class) & Class imbalance, Head-tail dichotomy & \cite{he2009learning,van2017devil,kang2019decoupling,wang2022imbalanced} & \cite{feldman2020does} \\
Meso (Sample) & Hard/atypical samples, Within-class variation & \cite{zhu2014capturing,carlini2019distribution,kishida2019empirical}& \cite{feldman2020neural,baldock2021deep,jiang2020exploring,garg2023memorization} \\
Micro (Noise) & (Label) Noise, Feature corruption & \cite{yi2019probabilistic,han2018co,zhang2018generalized,yu2018learning} &\cite{liu2020early} \\
\bottomrule
\end{tabular}
\end{table}

\begin{figure}[ht]
    \centering
    \includegraphics[width=0.8\textwidth]{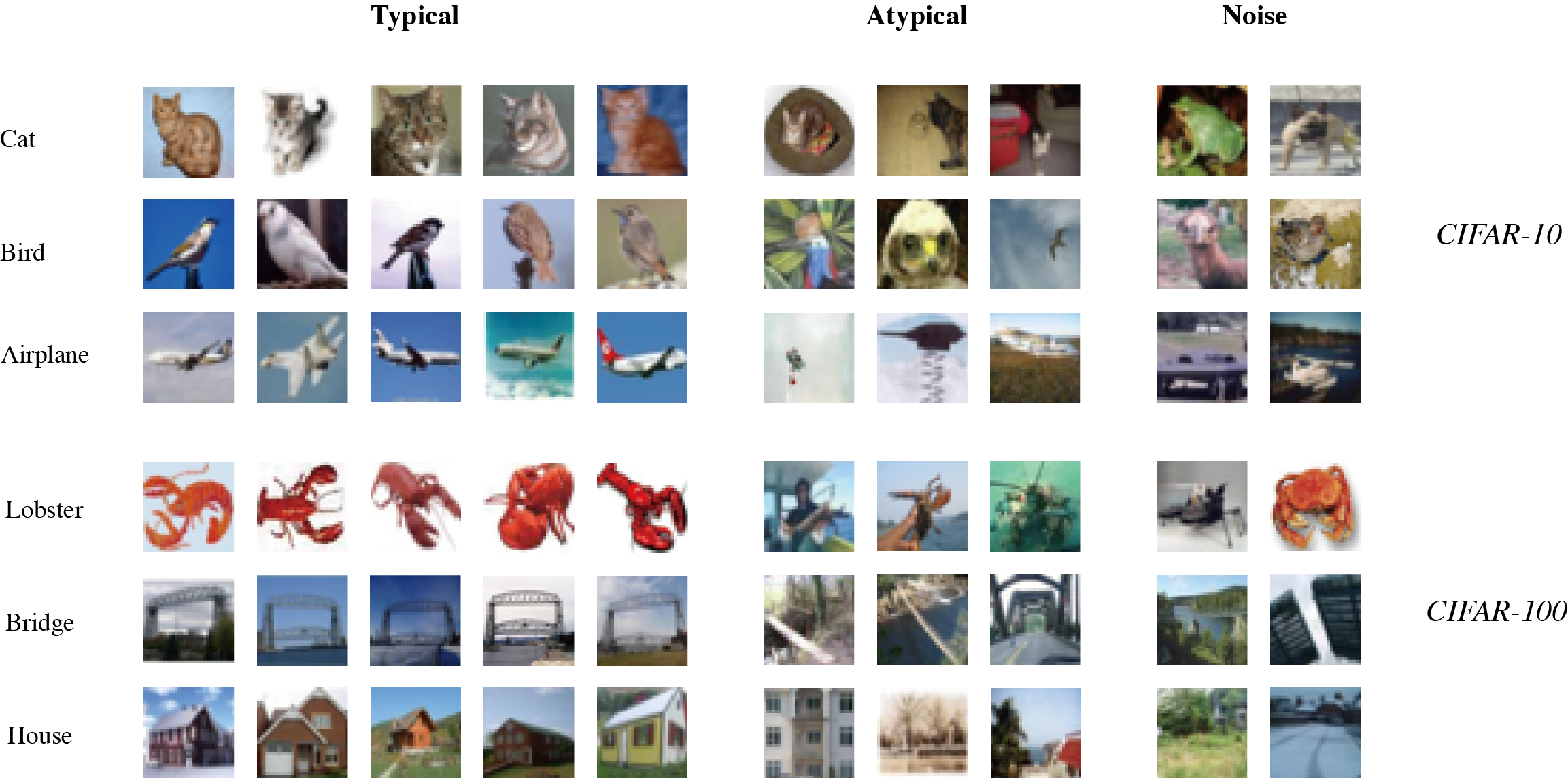}
    \caption{Sample images of different granularity from CIFAR-10 and CIFAR-100 datasets.}
    \label{fig:select_sample}
\end{figure}

\subsubsection{Three Granularity of Long-Tail Phenomena}

Long-tail phenomena manifest at distinct scales, where individual samples can be categorized as tailed data across three granularity, summarized in Table~\ref{tab:granularity} and examples are  illustrated in Fig.~\ref{fig:select_sample}:
\begin{itemize}
    \item \textbf{Imbalanced class}: Some classes (head) account for the majority of observations, while others (tail) are sparsely represented.
    This imbalance arises naturally in real-world data (e.g., medical imaging, rare species classification) or synthetically via sub-sampling. 
    \item \textbf{Atypical Sample (within-class)}:  Data points with rare or ambiguous features within their class, independent of class frequency. These reflect natural intra-class variation (e.g., in Fig.~\ref{fig:select_sample} the atypical samples in the "Lobster" class contain more background distractions compared to the typical ones. ). 
    \item \textbf{Noise (within-class)}: Instances with corrupted labels or features, often introduced during data collection (e.g., in Fig.~\ref{fig:select_sample}  the rightmost image depicts a crab but has been mislabeled as belonging to the "Lobster" class.). 
\end{itemize}


Most previous works on memorization have predominantly focused on addressing class imbalance.  While recent efforts have taken preliminary steps to disentangle class imbalance from the challenges posed by atypical samples, these advances remain nascent and narrowly focused. The field has yet to confront a deeper, under-appreciated granularity: the critical separation of atypical samples from noise. Though both manifest as ``hard'' cases in practice, they stem from fundamentally distinct mechanisms and demand divergent mitigation strategies. For instance, atypical samples, such as a lobster obscured by background clutter, require improved feature learning or targeted augmentation to enhance model robustness. In contrast, noise—like a crab mislabeled as a lobster—necessitates rigorous label correction or outlier suppression to prevent overfitting to spurious artifacts. Current theoretical frameworks often conflate these phenomena, treating noise as a subset of atypicality (e.g., Assumption \ref{asu:noise}). At the same time, empirical approaches misattribute their origins, applying augmentation to corrupted data or filtering rare-but-valid samples as outliers. This conflation can exacerbate, rather than resolve, long-tail challenges: models may overfit to noise while underfitting true intra-class diversity, undermining reliability. By formalizing this distinction, we not only expose a gap in existing analyses, but also empower practitioners to diagnose and address long-tail phenomena with surgical precision, a prerequisite for building trustworthy ML systems that generalize robustly to real-world data's inherent complexity.

\subsection{Long Tail Distribution and Memorization}

Having introduced both long-tail distributions and data memorization, we now explore their interplay to address two key questions: 1) Is memorization avoidable in ML? 2) Is memorization always harmful? And how does it relate to overfitting?

\subsubsection{Inevitability of Memorization} 
\citet{feldman2020does} theoretically demonstrates that, for classification problems, memorizing the \textbf{labels} of long-tailed data points is crucial to achieve \textbf{close-to-optimal generalization error}. The theoretical model proposed in~\citet{feldman2020does} has been empirically validated in the context of image classification by \citet{feldman2020neural} and further corroborated by \citet{zheng2022empirical} through studies in natural language processing (NLP). These findings collectively highlight the critical role of memorization in learning from long-tailed data distributions. 
Notably, the theoretical model assumes a specific data distribution,
summarized as follows:
\begin{assumption}\label{asu:subp}
[Low-Frequency Subpopulations]
Low-frequency subpopulations occupy a relatively significant portion of the data distribution (e.g., 5–10\% of the dataset).  Certain test examples depend heavily on corresponding training examples within these subpopulations. Removing or ignoring these subpopulations would result in a substantial drop in accuracy for the dependent test examples, while leaving the performance on unrelated examples unaffected.
\end{assumption}
\begin{assumption} \label{asu:noise}
[Indistinguishability of Subpopulations and Low Noise Rate]
Within subpopulations, the algorithm cannot reliably distinguish true atypical examples from noise or outliers. Consequently, algorithms tend to memorize both types of data. Generalization performance improves when the noise rate is low. However, if the noise rate is high, memorization of erroneous patterns may degrade model performance.
\end{assumption}

In addition to the theoretical analysis via an optimization perspective, \citet{brown2021memorization} provide an information-theoretical foundation for the necessity of memorization in long-tailed distributions. Using mutual information \citep{cover2012elements} as a framework, they show that memorizing singleton data points (subgroups containing only one data point) is critical for achieving high-accuracy learning in such tasks.
\begin{theorem}
For certain natural tasks \( q \), every dataset \( X \) contains a subset \( X_S \) (singleton data) such that \( X_S \) has an expected size of \( \Omega(n) \), where \( n \) is the dataset size. The entropy of \( X_S \), conditioned on the data distribution \( P \), denoted as \( H(X_S | P) \), scaled as \( \Omega(nd) \), where \( d \) is the data's dimensionality. 

Any learning algorithm \( A \) achieving an error close to the optimal error \( \text{err}_{q, n}(A_{\text{OPT}}) \) must satisfy:
    \[
    I(X_S; M | P) = (1 - o(1)) H(X_S | P),
    \]
    where \( M \) represents the trained model, and \( I(X_S; M | P) \) is the mutual information between \( X_S \) and \( M \), conditioned on \( P \).

\end{theorem}

This theorem formalizes the idea that singleton data points, which lack shared patterns with other examples, must be explicitly memorized for optimal learning. 
However, the analysis assumes extreme data scenarios in which singleton data points play a central role. Although these settings emphasize the theoretical necessity of memorization, their generalizability to more typical datasets is unclear. In particular, it is uncertain how the results extend to cases where subgroups contain more than one data point, which is a more realistic setting in many practical applications.

    
    

\subsubsection{Memorization vs. Overfitting} 
Memorization and overfitting are related yet distinct concepts. While memorization involves retaining specific examples, overfitting implies a model’s inability to generalize beyond the training data. 
The impact of memorization is highly context-dependent. \citet{zhang2017rethinking} show that over-parameterized neural networks can memorize random labels yet still generalize well on structured, real-world data, challenging the traditional view that memorization inherently leads to overfitting. Their follow-up work \cite{zhang2021understanding} further emphasizes that memorization of all training samples, including noisy ones, is not always at odds with generalization \cite{belkin2018overfitting,belkin2019reconciling,ma2018power,feldman2020does}. Conversely, excessive memorization of noisy or redundant data can lead to overfitting. \citet{chatterjee2020making, zielinski2020weak,chatterjee2022generalization} examine memorization by analyzing the directions of gradients of each sample and showed that weak (less dominanted) directions are responsible for memorizing hard and mislabeled samples, thereby resulting in overfitting. They further show that suppressing these weak directions improves generalization.
These underscore the nuanced relationship between memorization and overfitting: memorization is not inherently harmful but dependent on factors such as data distribution, noise levels, and training dynamics.

In summary, memorization is often unavoidable in ML, particularly when  handling rare examples in long-tailed distributions. 
Although it can sometimes lead to overfitting, it is not an invariable outcome in every case.
Therefore, striking a balance between necessary memorization and mitigating overfitting is essential for building trustworthy ML systems.
\section{Impacts of Long-Tail Data at Different Levels of Granularity}
\label{sec:granular}

The interplay between long-tail distributions and memorization manifests across three granularity levels, each presenting distinct challenges and mitigation strategies. This hierarchical perspective helps disentangle when memorization becomes necessary, harmful, or indicative of deeper data quality issues.

\subsection{Macro-Level: Class Imbalance and Bias}
\label{ssec:macro}

The memorization perspective reveals fundamental tensions in class-imbalanced learning. Training on long-tailed distributions causes models to excel on majority classes while underperforming on tail classes \cite{he2009learning,buda2018systematic,wang2022imbalanced}. This occurs because majority samples provide sufficient features for generalization, while tail classes force memorization of scarce patterns.  Privacy leakage risks vary across different subclasses, with smaller subclasses being more atypical to the model due to their under-representation during training, which makes them more vulnerable to attacks, such as MIA \citep{kulynych2019disparate}. 

From the memorization perspective, three key strategies are proposed to address this imbalance:
\begin{itemize}
\item \textbf{Re-sampling}:
Enforcing balanced class representation in training batches - by oversampling tail classes or undersampling head classes - directly counter the memorization bias toward majority classes~\citep{shen2016relay};

\item \textbf{Re-weighting}: Adjusting loss contributions via class-specific weights, effectively reducing the influence of frequently encountered head class samples while amplifying tail class signals, which reorients the model's memorization focus toward underrepresented classes \citep{cui2019class,tan2020equalization};

\item \textbf{Focal Loss}: Implementing sample level memorization control by automatically up-weighting hard examples, many of which reside in tail classes due to their inherent scarcity \citep{ross2017focal}.
\end{itemize}

However, all the three approaches face inherent trade-offs - minimizing majority-class memorization can underfit valid patterns, while over-emphasizing tail classes may lead to overfitting noises.
The tensions are particularly intensified in adversarial settings, where robustness gains concentrate on majority classes at the tail's expense \citep{wu2021adversarial}. Recent solutions adapt re-weighting principles to robustness objectives~\citep{li2023alleviating,yue2024revisiting}, however, fundamental challenges persist in balancing memorization across classes.

\subsection{Meso-Level: Atypical Samples and Their Cascading Risks}
\label{ssec:meso}

At the sample level, memorization manifests most acutely in atypical instances. The privacy and security implications are severe and multifaceted:
\begin{itemize}
\item \textbf{Differential Privacy Costs}: Differentially Private Stochastic Gradient Descent (DP-SGD) \cite{abadi2016deep} disproportionately degrades the performance accuracy of atypical samples by attenuating their gradient signals \cite{carlini2019distribution}, with longer tails exacerbating privacy-utility trade-offs \cite{suriyakumar2021chasing}.
\item \textbf{Membership Inference Vulnerabilities}: Similar to the class imbalance case, atypical samples are also more volunerable to MIAs due to their distinct memorization signatures \cite{li2024privacy,luo2024demem}.
\item \textbf{Adversarial Exploitation}: Attackers target atypical samples through backdoor insertion \cite{gao2023not} and data poisoning \cite{tramer2022truth}.  For example, \citet{gao2023not} state that  'not all samples are created equal', namely the difficulty of backdoor attacks to samples are different and DNNs have different learning abilities for different samples. More specifically, it proposes to deploy three classical difficulty metrics, including 1) loss value, 2) gradient norm, and 3) forgetting event, for selecting hard samples. \citet{tramer2022truth} demonstrate that, by exploiting the long-tail distribution and the memorization property of models, adversaries can conduct highly effective data poisoning. Specifically, they show that poisoning a small fraction ($< 0.1$\%) of a training dataset can significantly boost the effectiveness of inference attacks, including membership, attribute, and data extraction.  
\end{itemize}
The meso-level analysis underscores a critical insight: atypical samples function as memorization amplifiers, setting off cascading vulnerabilities that propagate to privacy breaches and adversarial attacks.

\subsection{Micro-Level: Noise Samples}
\label{ssec:micro}
Previous research shows that deep neural networks tend to memorize noisy labels during training, which results in a decline in generalization~\citep{zhang2017rethinking,arpit2017closer,liu2020early}. This issue has been extensively studied; for a detailed overview, please refer to the comprehensive survey by \citet{algan2021image}, which categorizes the proposed algorithms into two main groups: noise model-based methods (e.g. label correction \citep{yi2019probabilistic} and sample selection \citep{han2018co}), focusing on estimating the noise structure to mitigate its impact, and noise model-free methods, aiming to develop inherently noise-robust algorithms through approaches such as robust losses \cite{zhang2018generalized}, or alternative learning paradigms \cite{yu2018learning}.

\section{Relevance of Memorization in Trustworthy AI}

Trustworthy ML attributes often interact in complex ways, and analyzing their relationships is essential for building reliable and ethical ML systems. Memorization offers a valuable perspective for understanding how these attributes are interconnected and how they influence model behavior. 
\subsection{Memorization and Privacy}

\paragraph{Provable Privacy}  
There is a fundamental connection between memorization score and DP. Recall the definition of DP \citep{dwork2006,dwork2006calibrating,dwork2009differential}:  
A randomized algorithm $\mathcal{A}$ satisfies $(\epsilon, \delta)$-DP if for any two neighboring datasets $D_{\text{tr}}$ and $D_{\text{tr} \setminus (x,y)}$ differing by a single sample $(x,y)$, and for all measurable subsets $S$:

\[
\Pr[\mathcal{A}(D_{\text{tr}}) \in S] \leq e^{\epsilon} \Pr[\mathcal{A}(D_{\text{tr} \setminus (x,y)}) \in S] + \delta.
\]

Applying this to the definition of label memorization score in Eq. \eqref{eq.labelm} and using the approximation $e^\epsilon \approx 1 + \epsilon$ for small $\epsilon$, we obtain:

\[
\text{mem}(\mathcal{A}, D_{\text{tr}}, (x,y)) \leq \epsilon \Pr_{f_\theta \sim \mathcal{A}(D_{\text{tr} \setminus (x,y))}} [ f_\theta (x) = y ] + \delta.
\]

Thus, the memorization score is at most $\mathcal{O}(\epsilon)$, meaning DP constraints significantly limit the degree to which individual samples influence model predictions. This result highlights that memorization fundamentally captures the privacy risk under DP ~\citep{feldman2020does, li2024privacy}, which is widely regarded as a strong privacy guarantee.

\paragraph{Empirical Privacy Attacks}  Memorization has motivated the design of new powerful privacy attacks. 
It has been criticized in several works that the previous MIAs often have quite high false positive rates (FPR), i.e., many non-members are falsely identified as members~\citep{rezaei2021difficulty, carlini2022membership, hintersdorftrust}. However, attacks should obtain meaningful attack rates with low FPR, as it is more realistic for practical applications such as computer security \citep{kolter2006learning,lazarevic2003comparative}.  One of the reasons is that the attack results of many traditional MIAs  are not consistent with the distribution of memorization scores, thereby producing misleading results~\citep{li2024privacy}. On the contrary,  MIAs that exhibit a high degree of consistency with memorization scores should effectively capture how a model's behavior varies when a specific sample is included or excluded from the training set. In this context, we highlight three example MIAs that leverage such behavioral differences: the Bayes calibrated loss approach \citep{sablayrolles2019white}, the difficulty calibrated loss approach \citep{watson2021importance}, and Likelihood Ratio Attack (LiRA) \citep{carlini2022membership}. Note that LiRA is essentially an evolved form of the earlier two methods. Hence, MIAs that better capture the notion of memorization produce more reasonable attack results, i.e., obtaining meaningful attack rates under low FPR regions. 

\citet{carlini2022privacy} introduce the Privacy Onion Effect and empirically show that removing these high-risk outlier data points to mitigate privacy vulnerabilities inadvertently exposes previously safe inlier data points to similar risks, creating a cascading effect of vulnerability.
This phenomenon  emphasizes the iterative and relative nature of memorization and highlights the limitations of simple data removal strategies for ensuring model privacy.


\subsection{Memorization and Adversarial Robustness}

Memorizing mislabeled noise hurts the model's adversarial robustness \cite{sanyal2020benign}. It empirically shows that training samples misclassified by adversarially trained models but correctly classified by naturally trained models exhibit higher self-influence (reflecting greater atypicality; cf. Section \ref{subsec:proxies} for a detailed explanation of self-influence and its computation) compared to the overall self-influence distribution across the entire training dataset. In other words, the AT excludes fitting these samples which are hard to memorize \citep{sanyal2020benign}. \citet{xu2021towards} further explore the relationship between adversarial robustness and memorized samples, identifying two types of atypical samples: benign and harmful. They observe that memorizing benign atypical samples can effectively enhance a DNN's accuracy on clean data, but has minimal impact on improving adversarial robustness. In contrast, memorizing harmful atypical samples can degrade the DNN's performance on typical samples. To address this issue, they propose Benign Adversarial Training (BAT) \citep{xu2021towards}, a method designed to guide AT by minimizing the influence of harmful atypical samples while prioritizing the fitting of benign atypical samples.

Previous studies have demonstrated that DNNs possess sufficient capacity to easily memorize data, even when the labels are random or the inputs are pure noise \citep{zhang2017rethinking}. Building on this, \citet{dong2021exploring} empirically shows that DNNs can similarly memorize adversarial examples with completely random labels, although the convergence behavior varies depending on the used AT algorithm. Furthermore, they identify a critical drawback of memorization in AT that it could result in robust overfitting \citep{rice2020overfitting}.

However, it has also been empirically proved that improving robustness in ML models, particularly through AT, can inadvertently increase privacy risks \citep{song2019privacy,li2024privacy}. Specifically, AT makes models more susceptible to MIAs. From the memorization perspective, one explanation for this heightened privacy leakage is that adversarially trained models tend to memorize more training samples than non-adversarially trained models \citep{li2024privacy}. To address this issue, \citet{luo2024demem} propose a method to mitigate the privacy leakage risks of AT models by reducing the influence of high-risk samples, i.e., samples with high memorization scores, while preserving both robustness and natural utility.

\subsection{Memorization and Fairness}
\citet{chang2021privacy} demonstrate that fairness-driven algorithms \citep{agarwal2018reduction}, designed to balance error rates across groups that are defined by protected attributes, often rely on memorizing data from under-represented subgroups. While promoting equitable outcomes, this approach inadvertently heightens the risk of information leakage concerning these unprivileged groups, raising critical concerns about privacy and security in fairness-sensitive contexts.

\citet{you2025silent} investigate fairness issues in neural networks by exploring the connection between imbalanced group performance and memorization. The study identifies that spurious memorization, confined to a small subset of ``critical neurons'', leads to disproportionately high accuracy for majority groups while minority groups suffer from reduced test accuracy, despite achieving comparable training accuracy. The authors demonstrate that these critical neurons memorize minority group examples using spurious features, resulting in imbalanced performance. To address this issue, they intervene with the model by pruning these neurons, showing that spurious memorization can be mitigated, improving minority group accuracy with minimal impact on majority groups. This work highlights spurious memorization as a fundamental cause of fairness issues, providing a novel perspective on addressing imbalanced performance in neural networks.

\subsection{Memorization and Transparency \& Explainability}
Analyzing memorization mechanisms helps to enhance the explainability and transparency of ML models.




\paragraph{Temporal Dynamics of Memorization}
The timing of memorization during training follows distinct phases. \citet{arpit2017closer} find that DNNs prioritize learning simple and generalizable patterns in the early stages of training. For example, during this phase, a model trained on image data might focus on shared features like edges and textures, or syntactic rules in NLP tasks. Memorization typically emerges later in the training process as the model begins to overfit to noise, outliers, or complex patterns that cannot be generalized. This shift from generalization to memorization is influenced by factors such as data complexity, model capacity, and regularization strategies. 
Additionally, \citet{liu2021understanding} explore memorization from an optimization perspective, finding that DNNs simultaneously optimize easy and difficult examples, with easy ones converging faster. This concurrent optimization underscores the need for strategies that balance learning across different data complexities to mitigate overfitting. \citet{toneva2019empirical} investigate the concept of "forgettable examples" in DNN training. They find that certain difficult or atypical examples are memorized and forgotten multiple times during training. This finding suggests that memorization is not a single event but a dynamic process, where some examples require repeated exposure to be encoded. \citet{carlini2023extracting} examine memorization in diffusion models and observed that specific instances (e.g., rare or sensitive images) are reproduced only after extended training. This indicates that memorization is a late-stage phenomenon in generative models, often occurring after the model has learned the general structure of the data distribution.  \citet{mangalam2019deep} show that DNNs naturally adopt an ``easy-to-hard'' curriculum during training, where the model first learns simple patterns that explain the majority of the data. Memorization of harder or noisier examples occurs progressively as the training continues. 

\paragraph{Spatial Localization of Memorization}  The localization of memorization in DNNs, whether confined to specific layers or distributed across the network, remains inconclusive as it varies on the task and architecture. In computer vision, studies have shown that earlier layers tend to generalize, extracting broad and reusable features, while deeper layers are more prone to memorization. \citet{baldock2021deep} observe that deeper layers adapt to finer details, leading to memorization, while \citet{stephenson2021geometry} explain this behavior as a result of geometric transformations in data representation. In contrast, \citet{maini2023can} argue that memorization is not isolated to deeper layers but is distributed across multiple layers throughout the network. In NLP tasks, \citet{dankers2024generalisation} reveal that memorization is a gradual and task-dependent process. Their findings challenge the ``generalization-first, memorization-second'' hypothesis, showing that early layers are often crucial for encoding task-specific patterns necessary for memorization, while deeper layers refine and contextualize these representations.

Hence, current insights into memorization's temporal and spatial patterns remain largely observational and task-dependent, with no consensus on how best to regulate them for better trustworthiness. Further investigation is needed to clarify memorization’s role in trustworthy ML, e.g., whether adversarial attacks can exploit memorized layers to undermine robustness. Addressing these gaps will be essential for balancing model utility and societal trust.


\begin{figure}[ht]
    \centering
    \includegraphics[width=0.9\textwidth]{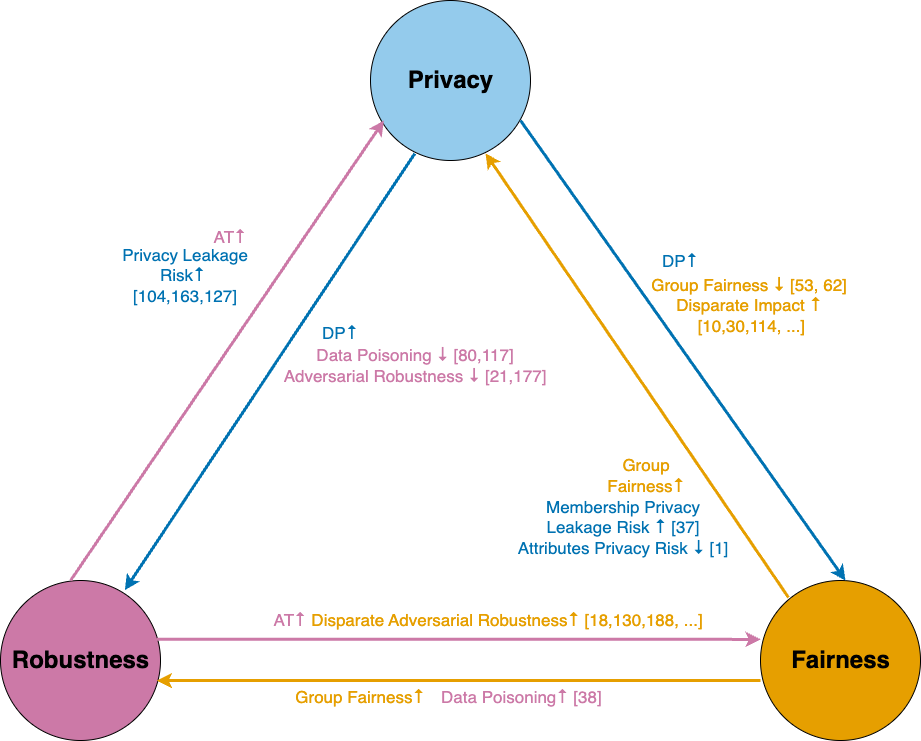}
    \caption{An overview of trustworthy attribute trade-off, providing an intuitive trade-off guide to the detailed information in Table \ref{tab:simplified-defences}.}
    \label{fig:interaction}
\end{figure}

\begin{table}[h!]
\centering
\renewcommand{\arraystretch}{1.3}
\begin{tabular}{|p{5cm}|p{2.5cm}|p{2.5cm}|p{1.7cm}|p{1.7cm}|}  
\hline
\textbf{Trustworthy Attribute} & \multicolumn{4}{|c|}{\textbf{Phenomenon}}\\
\cline{2-5}
& Data dependent & Mitigation & Inherent & Mitigation \\
\hline
\textbf{No Trustworthy Enhancement} &&&&\\ 
Generalization & $\mathbb{S}$\(\uparrow\) \cite{feldman2020does,feldman2020neural,zheng2022empirical} &&& \\
&$\mathbb{N}$\(\downarrow\) \cite{zhang2017rethinking,arpit2017closer,liu2020early}& $\mathbb{N}$ \cite{algan2021image}&&\\ 
$\mathtt{F}$ •  Disparate Generalization & $\mathbb{S}$ \cite{you2025silent}&$\mathbb{S}$ \cite{ross2017focal}&&\\ 
&$\mathbb{C}$ \cite{he2009learning,buda2018systematic,wang2022imbalanced} &$\mathbb{C}$ \cite{shen2016relay,cui2019class,tan2020equalization}&& \\ 
$\mathtt{P \& F}$ • Disparate Privacy Leakage Risk & $\mathbb{S}$ \cite{li2024privacy,luo2024demem}&&  \cite{shokri2017membership}&\\
&$\mathbb{C}$ \cite{kulynych2019disparate}&&&\\
\hline
\textbf{Robustness (AT \(\uparrow\))}&&&&\\ 
Generalization &&&\(\downarrow\) \cite{tsipras2018robustness,zhang2019theoretically}&\cite{zhang2019theoretically,wu2020adversarial}\\

$\mathtt{P}$ • Membership Privacy Leakage Risk & && \(\uparrow\) \cite{song2019privacy,li2024privacy} &  \cite{luo2024demem} \\ 
$\mathtt{P}$ • Reconstruct Privacy Leakage Risk &&&  \(\uparrow\) \cite{mejia2019robust}&\\ 
$\mathtt{R\& F}$ • Disparate Adversarial Robustness &$\mathbb{C}$\(\uparrow\) 
 \cite{wu2021adversarial,wang2022imbalanced}&$\mathbb{C}$ \cite{wu2021adversarial,li2023alleviating,wang2022imbalanced,yue2024revisiting} & \(\uparrow\) \cite{benz2021robustness,nanda2021fairness}&  \cite{xu2021robust,ma2022tradeoff} \\
$\mathtt{R}$ • Adversarial Robustness & $\mathbb{N}$\(\downarrow\) \cite{sanyal2020benign,xu2021towards} &&&\\ 
$\mathtt{R}$ • Data Poisoning & && \(\downarrow\) \cite{tao2021better,fu2022robust}&\\

\hline
\textbf{Privacy (DP \(\uparrow\))}&&&&\\ 
Generalization &&&\(\downarrow\) \cite{abadi2016deep}&\\
$\mathtt{F}$ • Disparate Generalization & $\mathbb{S}$\(\uparrow\) \cite{carlini2019distribution} &&&\\ 
&$\mathbb{C}$\(\uparrow\) \cite{bagdasaryan2019differential,suriyakumar2021chasing,zhong2022understanding} &$\mathbb{C}$ \cite{xu2021removing,uniyal2021dp}&&\\ 
$\mathtt{F}$ • Disparate Defense & $\mathbb{S}$ \cite{luo2024demem} &$\mathbb{C}$ \cite{zhong2022understanding} &&\\ 
$\mathtt{F}$ • Group Fairness & $\mathbb{C}$\(\downarrow\) \cite{farrand2020neither} &$\mathbb{C}$ \cite{tran2021differentially,khalili2021improving,pannekoek2021investigating} & \(\downarrow\) \cite{cummings2019compatibility} &  \cite{cummings2019compatibility}\\ 
$\mathtt{R}$ • Adversarial Robustness & && \(\downarrow\) \cite{tursynbek2020robustness,boenisch2021gradient} &\\ 
$\mathtt{R}$ • Data Poisoning & && \(\downarrow\) \cite{ma2019data,hong2020effectiveness}&\\
\hline
\textbf{Fairness (Group Fairness \(\uparrow\))}&&&&\\
Generalization & $\mathbb{N}$\(\downarrow\)\cite{roh2021sample}&$\mathbb{N}$\cite{roh2021sample}&\(\downarrow\)\cite{zhao2022inherent}&\\

$\mathtt{P}$ • Membership Privacy Leakage Risk &&&  \(\uparrow\) \cite{chang2021privacy} & \(\downarrow\) \cite{tian2024fairness}\\ 
$\mathtt{P \& F}$ • Disparate Membership Privacy Leakage Risk & $\mathbb{C}$\(\uparrow\) \cite{chang2021privacy} &&&\\ 
$\mathtt{P}$ • Attributes Privacy Risk &&&  \(\downarrow\) \cite{aalmoes2025alignment} &\\
$\mathtt{R}$ • Data Poisoning &&&  \(\uparrow\) \cite{chang2020adversarial} &\\

\hline
\end{tabular}
\caption{The relationships between variations in different trustworthy ML attributes and their corresponding effects on other attributes. $\mathtt{P}$: Privacy attribute,$\mathtt{R}$: Robustness attribute,$\mathtt{F}$: Fairness attribute; $\mathbb{C}$: Imbalanced class, $\mathbb{S}$: Atypical sample, $\mathbb{N}$: Noise; \(\uparrow\): Increase, \(\downarrow\): Decrease}
\label{tab:simplified-defences}
\end{table}

\section{Interactions of Trustworthy ML Attributes via Memorization}
Memorization is deeply rooted in ML due to the inherent challenge of learning from long-tail data distributions, where rare or atypical samples necessitate memorization to achieve generalization. This necessity creates inherent tensions between key attributes of trustworthy ML: privacy mechanisms (e.g., DP) that suppress memorization to protect sensitive data risk discarding critical atypical samples required for fairness, while robustness strategies (e.g., AT) may amplify bias by discarding memorized minority-class patterns. These trade-offs underscore the interconnectedness of trustworthiness attributes — interventions enhancing one aspect (e.g., privacy, robustness, or fairness) often inadvertently compromise the others. Table \ref{tab:simplified-defences} summarizes these interactions, emphasizing the need for holistic approaches to regulate memorization while balancing competing objectives.




\subsection{Impact of Privacy Enhancement}
Differential privacy has proven to be effective in countering privacy risks posed by memorization \citep{carlini2022membership,luo2024demem}, however, it can exacerbate generalization performance disparities, particularly impacting underrepresented groups \citep{bagdasaryan2019differential}. The classes with fewer samples, which are already underrepresented and have lower accuracy, end up losing even more utility by implementing the DP-SGD -  ``The poor become poorer''. This indicates that privacy enhancement may come at the cost of increased unfairness for complex and minority subgroups. \citet{farrand2020neither} demonstrate that even minor imbalances in class samples and weak privacy guarantees can lead to disparate impacts. They also employ group fairness metrics to show that increasing privacy protection may, in turn, compromise group fairness.

The fundamental incompatibility between strict DP guarantees and group fairness metrics is formally established by \citet{cummings2019compatibility}. Their work proves that while exact fairness under DP constraints is theoretically unattainable, a practical reconciliation can be attained through the Exponential Mechanism. This approach enables alpha discrimination fairness — a relaxed fairness criterion — while maintaining DP guarantees, thereby establishing theoretical foundations for privacy-preserving learning under approximate fairness constraints.

To address generalization disparities in DP-trained models, \citet{xu2021removing} propose DPSGD-F, which implements group-specific adaptive gradient clipping. By dynamically adjusting clipping bounds according to the gradient distributions of distinct demographic groups, this method mitigates overfitting to majority groups while enhancing generalization for underrepresented classes.
Further, FairDP is proposed to utilize an adaptive clipping threshold to regulate the impact of instances in each class, enabling the model accuracy to be adjusted for classes based on its privacy cost and generalization disparities considerations \citep{tran2023fairdp}. 

\citet{uniyal2021dp} find that using Private Aggregation of Teacher Ensembles (PATE) \cite{papernot2016semi} to approach DP has less impact on performance bias.  
\citet{hong2020effectiveness} unifies the threat surface of poisoning attacks by identifying common gradient-level properties, proposes gradient shaping as a generic and attack-agnostic defense, and demonstrates the effectiveness of DP-SGD in mitigating various poisoning attacks \cite{jagielski2018manipulating,shafahi2018poison}, even with minimal privacy guarantees.\citet{ma2019data} demonstrate that DP provides protection against small-scale data poisoning attacks. However, as the scale of the poisoning increases, the effectiveness of this defense gradually diminishes. Furthermore, the effectiveness of the attack is inversely proportional to the privacy budget in DP.

Moreover, improving privacy through DP also poses challenges to robustness, as it has been shown in \cite{tursynbek2020robustness, boenisch2021gradient} that DP-trained models may become more susceptible to adversarial attacks and input perturbations compared to non-private counterparts. Tursynbek et al. \cite{tursynbek2020robustness} empirically analyzed the trade-off between privacy and robustness by evaluating DP models against adversarial attacks such as FGSM \cite{goodfellow2014explaining} and PGD \cite{mkadry2017towards}. Boenisch et al. \cite{boenisch2021gradient} extended this analysis by incorporating gradient-free \cite{brendel2017decision} and optimization-based \cite{carlini2017towards} adversarial attack methods. These studies reveal that DP models may be more vulnerable to adversarial attacks, as DP training components — such as gradient clipping and noise addition — can reduce robustness, particularly at higher noise levels.

\subsection{Impact of Robustness Enhancement}
\citet{tsipras2018robustness} empirically demonstrate that there is a trade-off between generalization and adversarial robustness arising from the features learned by standard training and robust classifiers are fundamentally different, a phenomenon that persists even with infinite data. \citet{zhang2019theoretically} further proved that the robust error can be tightly bounded by the sum of natural error (using a surrogate loss) and boundary error (measuring proximity to the decision boundary under adversarial perturbations). Based on this, they propose TRADES, a new adversarial defense method that optimizes a regularized surrogate loss, balancing natural accuracy and adversarial robustness. 

\citet{benz2021robustness} empirically studies robustness and inter-class fairness, showing that inter-class discrepancy persists across various setups and becomes more pronounced with AT, where enhancing robustness exacerbates fairness disparities. Attempts to mitigate this discrepancy by adjusting class-wise loss weights reveal that improving fairness for certain classes often decreases overall accuracy or robustness, highlighting the challenges of balancing these attributes. 

\citet{nanda2021fairness} introduces the phenomenon of robustness bias, which refers to the disparity in a model's susceptibility to perturbations across different data partitions, where some groups are significantly closer to or farther from the decision boundary, making them disproportionately vulnerable to adversarial attacks. 

\citet{tao2021better} demonstrates that AT effectively defends against delusive attacks \cite{newsome2006paragraph}, a form of data poisoning attack, by mitigating the detrimental effects of slight perturbations in training data, preventing models from relying on non-robust features. \citet{fu2022robust} further found that AT is effective against poisoning because it trains models on adversarial examples, which, despite being crafted from unlearnable data, still lead to higher loss and enable the model to learn from these examples, undermining the data poisoning.

There are some follow-up works \cite{xu2021robust,ma2022tradeoff} that focus on mitigating the unfairness issue of the AT model.

\citet{xu2021robust} introduced the Fair-Robust-Learning (FRL) framework, proposing two strategies to address robust fairness: FRL (Reweight), which adjusts class-wise loss weights by upweighting standard and boundary error costs for underperforming classes, and FRL (Remargin), which increases the adversarial perturbation margin for specific classes to reduce boundary errors and improve robustness. \citet{ma2022tradeoff} introduced the Fairly Adversarial Training (FAT) method, which incorporates the variance of class-wise accuracy as a penalty term within the loss function, effectively balancing the trade-off between robustness and robust fairness.

\subsection{Impact of Fairness Enhancement}
\citet{zhao2022inherent} theoretically proved that the tradeoff to satisfy group fairness requirements, the classifier must sacrifice accuracy for certain groups in order to enhance fairness for others.
\citet{tian2024fairness} improves group fairness through Fair Mixup, a data augmentation strategy that optimizes group fairness constraints by bridging sensitive groups with interpolated samples \cite{chuang2021fair}. They demonstrate that fairness-enhanced models offer better defense against MIA, showing that these models have a lower MIA attack success rate compared to biased models.
\citet{aalmoes2025alignment} find that ensuring group fairness also affords a protection against the attribute inference attack. With this group fairness restriction, it ensures that the model's predictions are independent of sensitive attributes.  
\citet{chang2020adversarial} empirically demonstrate a fundamental trade-off between fairness and data poisoning, showing that fairness constraint models are more vulnerable to adversarial poisoning attacks, which leads to sharp performance declines in test accuracy.
\citet{zeng2023adversarial} further extend this threaten scenario in the face attribute recognition task by designing a novel positioning method, which uses Maximum Mean Discrepancy (MMD) to perturb the data domain of the target demographic group, minimizing the distance to a reference group. Additionally, they apply Kullback-Leibler (KL) divergence to optimize instance-level perturbations, altering individual demographic attributes to induce bias without modifying labels. To address this issue, they proposed a robust-fair training to defend the potential fairness attack.
\citet{van2022poisoning} proposed framework formulates a data poisoning attack on fair ML as a bi-level optimization problem, where the attacker maximizes a weighted combination of accuracy and fairness loss by constructing a poisoning dataset. The framework includes three attack strategies to compromise the generalization and group fairness of the fairness-enhanced model: PFML-AS selects poisoning samples based on their impact on both accuracy and fairness; PFML-AF flips labels to disrupt model learning; and PFML-AM modifies features to alter decision boundaries.

The trade-offs between fairness, robustness, and privacy in ML are intricate. Current limitations underscore the necessity for further research in proposing integrated approaches to balance these factors to develop unified models that simultaneously optimize privacy, fairness, and robustness.



\section{A Case Study of Trustworthy Multilingual LLMs}
After presenting the existing results of trustworthy ML from a memorization perspective, we now dive into multilingual LLM to demonstrate how the memorization perspective serves as a fundamental way to interpret the trustworthiness of multilingual LLMs.

\subsection{Multilingual LLMs and Long-tail Low-Resource Languages}
Recent advancement in LLMs has significantly improved their cross-lingual capacities, particularly in comprehending multiple languages~\citep{conneau-etal-2020-unsupervised,conneau-etal-2018-xnli}, benefiting numerous NLP tasks~\citep{choi2021analyzingzeroshotcrosslingualtransfer, pikuliak2021cross}, especially with the recent GPTs~\citep{floridi2020gpt,achiam2023gpt}, LlaMas~\citep{touvron2023llamaopenefficientfoundation} and DeepSeek models~\citep{wu2024deepseek,liu2024deepseek}.
Training multilingual LLMs requires incorporating text data in a diverse range of languages. However, while increasing the number of languages potentially boosts downstream task performance for (extremely) low-resource languages, it would also undercut the performances of high-resource languages~\citep{chang2023multilingualitycurselanguagemodeling}. 
This is so-called \textit{curse of multilinguality}~\citep{conneau-etal-2020-unsupervised}, where there is a trade-off between the number of languages incorporated in pre-training LLMs and the performances of languages in NLP tasks, with the model capacity invariant.
While, there have been continuous endeavors to mitigate the curse of multilinguality in MLLMs~\citep{pfeiffer-etal-2022-lifting,chang2023multilingualitycurselanguagemodeling,blevins-etal-2024-breaking,blevins-etal-2024-breaking}, 
the inequality of language distribution in LLMs persists, which leaves a long-tail of under-represented languages. 
This disparity arises from various factors - ranging from unequal human resources and limited access to digital devices to the artifacts produced, including linguistic knowledge, data and technology, and also agency of community members - all of which are rooted in broader socio-political inequalities~\citep{nigatu-etal-2024-zenos}.
While there are over 7000 languages in the world, only a handful of them are well-represented in the Internet, and thereafter used for pre-training LLMs. 
In the sentinel work of~\citet{joshi-etal-2020-state}, the languages are taxonomized according to their ``richness'' in the digital age, in the context of data availability, with only seven languages such as widely used English, Spanish and Japanese being the obvious~\textbf{Winners}, then there are~\textbf{The Underdogs},~\textbf{The Rising Stars},~\textbf{The Hopefuls},~\textbf{The Scraping-Bys}, and eventually 2191 languages are classified as~\textbf{the Left-Behinds}.
While there is a problem of data scarcity for low-resource languages, there is a bigger problem of violating data sovereignty~\citep {walter2023indigenous} and over-exploitation of languages and their communities rather than embracing languages as a situated and embodied social practice~\citep{bird-2024-must}.
It is an essential and existential challenge of incorporating a large body of long-tailed low-resource languages in LLMs and consequently NLP research in general, while ensuring ethical operationalization and fairness for marginalized language communities. 
The following sections highlight the security and safety issues particularly for long-tailed low-resource languages in multilingual LLMs, illustrating a deeper layer of vulnerabilities.



\subsection{Multilingual LLM security and Long-tail Low-Resource Languages}

The burgeoning research in security and safety in multilingual LLMs has revealed an even bleaker situation for low-resource languages and their ripple effects in this space. Jailbreaking, techniques that bypass safety guardrails in LLM systems, often through adversarial prompts, to elicit harmful or restricted outputs ~\citep{gcg,li2024faster}, exacerbates these challenges.
It has been demonstrated that safety mechanisms are easier to bypass in lower-resource languages~\citep{yong2023low,deng2023multilingual,xu-etal-2024-cognitive,shen_language_2024},
with these languages producing more unsafe~\citep{deng2023multilingual} and irrelevant outputs~\citep{shen_language_2024}.
Prompts with a mixture of languages of different resource levels also prove more effective in jailbreaking the state-of-the-art LLMs than solely using high-resource languages~\citep{upadhayay_sandwich_2024,song_multilingual_2024}.
Moreover, multilingual backdoor attacks directly leverage the cross-lingual transferability to attack models by poisoning data in a low-resource language to affect the outputs in high-resource languages that are not poisoned, affecting both machine translation~\citep{wang-etal-2024-backdoor} and instruction tuning~\citep{he2024transferring}.   
More recently, textual embedding inversion attacks~\citep{song2020informationleakageembeddingmodels} have also emerged in multilingual settings, where the original texts are reconstructed from their embedding representations, revealing greater vulnerability in multilingual models than monolingual ones, especially for low-resource languages, such as Kazakh and Urdu~\citep{chen2024against,chen2024text}.

Although defense mechanisms have been explored to address and mitigate such vulnerabilities of multilingual LLMs, in pursuit of defending the LLMs, they either diminish LLMs' usefulness~\citep{deng2023multilingual}, or are ineffective in addressing the vulnerabilities, especially for low-resource languages~\citep{shen_language_2024}.
In summary, the inherent discrepancy of language distribution renders multilingual LLMs more vulnerable than monolingual ones, with 1) low-resource languages are particularly vulnerable in LLMs; 2) the vulnerability is transferable from low-resource to high-resource languages; 3) the defense mechanisms are largely ineffective to defend LLMs against multilingual attacks.

\subsection{Hallucinations in Multilingual LLMs and Long-tail Low-Resource Languages}

While LLMs have demonstrated remarkable capabilities in natural language understanding and text generation, and a broad range of NLP tasks, LLMs have a tendency to exhibit hallucinations, where the generated content is contradictory or counterfactual. 
The problem of hallucinations in LLMs not only significantly undermines their reliability and trustworthiness, but further erodes human knowledge with propagating falsities, particularly as education becomes more reliant on LLMs.
Therefore, the research on mitigating and guardrailing LLMs against hallucinations is essential to trustworthy ML.
The works on hallucinations have been prolific, however, they mostly focus on English contexts, ~\citep{zhang2023siren}.


More recently,~\citet{guerreiro2023hallucinations} conduct a thorough analysis over one hundred language pairs across resource levels on both neural machine translation models and GPT LLMs, highlighting that models struggle mainly in low-resource languages with Hallucinations, and specifically, they may reveal toxic patterns that can be traced back to the training data.
The vulnerability in low-resource languages is further highlighted in the multilingual summarization task, where hallucination in terms of faithfulness to the original document is worse in low-resource languages when summarization requires cross-lingual transfer~\citep{qiu2023detecting}.
Moreover, an examination of 19 languages at different resource levels at various resource levels, in free-form text generation task, reveals a notable factuality gap, measured by \textsc{factscore}, between low-resource and high-resource languages, with hallucination proving significantly more severe in the former~\citep{chataigner2024multilingual}.
To assist a more thorough evaluation of LLM robustness in retrieval augmented generation,~\citet{thakur2024knowing} curate a human-annotated dataset~\textbf{NoMIRACL} across 18 languages. 
Although most LLMs produce a high percentage of non-relevant information across languages, ironically, low-resource languages like Swahili and Yoruba exhibit the lowest hallucination rates, likely because their limited resources lead them to frequently respond with ``I don’t know.''
A further effort in multilingual space in LLM hallucination research is~\citet{islam2025llmshallucinatelanguagesmultilingual}. 
The authors train a multilingual hallucination detection model and conduct a study across 30 languages and 6 LLM families, however, finding that there is no notable correlation between hallucination rates and the language resource level.
In comparison, \citet{kang2024comparinghallucinationdetectionmetrics} assess and compare various factual hallucination metrics in 19 languages across resource levels and whether they align with human judgment, revealing a significantly diminished recall and increased hallucinations, especially in low-resource languages, and that the current existing automatic factuality metrics are flawed to evaluate non-English languages. 

To mitigate hallucinations in various NLP tasks, specific strategies have been adopted accordingly.
Employing more diverse models, trained on different data or with different procedures, is used to mitigate hallucination in neural machine translation systems~\citep{guerreiro2023hallucinations}.
To reduce hallucinations in cross-lingual summarization, loss weighting of each training data by a novel faithfulness score is adopted~\citep{qiu2023detecting}.
Overall, research on hallucinations in LLMs and multilinguality, especially in long-tailed low-resource languages, is still in the early stages. More severely, without proper metrics to evaluate hallucinations in multiple languages rather than English, it is hard to assess the scale and the scope of hallucinations in low-resource languages in LLMs.


\subsection{Long-tail Low-Resource Languages and Memorization}


Multilingual LLMs consist of natural long-tail data, i.e., low-resource languages.
Recent work in the intersection of long-tailed low-resource languages and memorization reveals that for low-resource or extremely low-resource languages, the 
machine-translation model becomes a retriever of the contents of the training data, and further question the current metrics such as BLEU score in machine translation, which would not discover the memorization issue in this case~\citep{cavalin2024fixing}.
Moreover, to measure the phenomenon language confusion in LLMs, where the LLMs output texts in languages that are erroneous and unintended, ~\citet{chen2025largelanguagemodelseasily} propose a re-weighted entropy metric to highlight the long-tail of low-resource languages, and find consistent patterns of language confusion across LLMs and datasets, such as that LLMs are more prone to produce erroneous languages when the source languages are low-resource languages such as Indonesian and Hindi in comparison to high-resource languages such as French and Spanish.

Overall, the intersection of multilingual LLMs, long-tailed low-resource languages, and memorization is extremely under-explored. 
As shown in~\citet{chen2025largelanguagemodelseasily}, language confusion patterns in LLMs - revealed by emphasizing long-tailed language distribution - correlate with language similarities, which can be exploited for cross-lingual attacks in a more targeted manner.
Moreover, illustrated in~\citet{satvaty2024undesirable}, investigating whether a low-resource language setting is more prone to memorization in comparison to higher-resource languages, can be essential not only for multilingual LLM privacy and data leakage but also for ensuring AI safety in communities using multilingual LLMs for low-resource languages.
Further research in this directions can provide meaningful insights for safer, more secure and trustworthy ML applications.

As demonstrated in this section, current research and development of LLMs pose unique and challenging security and safety issues to long-tailed low-resource languages, especially when the LLMs are not designed to guardrail the risks for these long-tailed languages, mainly due to the focus of NLP communities being on the few high-resource languages. 
Nevertheless, this area is ever-evolving and presents numerous open revenues for future research.

\section{Trending and Open Problems}
\subsection{Theoretical Gap of Memorization in ML and trustworthy ML}
Existing theoretical work has established that memorizing long-tailed or atypical samples is necessary to achieve optimal generalization \citep{feldman2020does, brown2021memorization}. However, these analyses rely on idealized assumptions, such as the indistinguishability of atypical samples and noise (Assumption \ref{asu:noise}), which may not hold in practice. While memorization is well-studied in the context of generalization, its fundamental role in trustworthy ML, particularly its interplay with adversarial robustness, fairness, and transparency,  remains under-explored.

For instance, \citet{d2021tale} demonstrate that memorizing atypical samples improves generalization but degrades adversarial robustness, suggesting an inherent tension between utility and trustworthiness. This raises critical questions: 
Is memorization still necessary for trustworthy ML, or does it inherently conflict with objectives like robustness and fairness? 
If memorization is unavoidable, what granularity (noise, atypicality, class imbalance) should models prioritize memorizing to balance competing trustworthiness goals? These open questions underscore the urgent need for rigorous theoretical investigation into memorization’s dual role in trustworthy ML, particularly under real-world conditions where generalization and societal trust must co-exist.

\subsection{Dual Roles of Memorization across Attributes}
While memorizing tailed data has proven beneficial for improving model generalization \citep{feldman2020does,brown2021memorization}, its role in trustworthy ML is far more nuanced. Memorization occupies a dual role: helpful memorization (e.g., retaining rare, high-utility patterns like medically critical outliers) enhances accuracy, while harmful memorization (e.g., overfitting to noise or spurious correlations) undermines trustworthiness. For instance, \citet{d2021tale} expose a critical paradox: memorizing atypical samples boosts generalization on underrepresented classes yet erodes adversarial robustness. This tension reveals a broader, understudied challenge—the context-dependent trade-offs of memorization across trustworthiness attributes like robustness, fairness, and privacy.




\subsection{Sensitivity or Consistency of Memorization Scores}
A critical open challenge is the sensitivity of memorization scores to model architectures, training methodologies (e.g., regularization), and data distributions. 
For example, \citet{feldman2020neural} compares memorization and influence estimates across different neural network architectures (ResNet series, Inception, and DenseNet100) on CIFAR-100 \citep{krizhevsky2009learning} by assessing Jaccard similarity and average estimation differences. They reveal that the consistency of memorization estimates across different neural network architectures is high with minimal selection bias, while influence estimates show greater sensitivity to threshold choices, with performance closely tied to model accuracy.

Recent work by \citet{kwok2024dataset} further investigates sensitivity through the lens of example difficulty scores, a proxy for memorization (cf. Section \ref{subsec:proxies}). Their study reveals that metrics like VoG, AUM, and PD exhibit high run-to-run variance, especially for hard examples, making threshold-based rankings unreliable with limited runs. Notably, based on the top eight most and least sensitive CIFAR-10 examples ranked by changes in difficulty scores across different model architectures, as shown in \citet{kwok2024dataset}. We observe that the most sensitive samples to model architecture differences (e.g., ResNet vs. VGG) tend to be highly atypical, whereas the most model-independent examples in terms of difficulty scores are those that appear the most typical. These findings align with our earlier discussion: memorization proxies are inconsistently aligned with atypicality/noise distinctions, raising concerns about their applicability in trustworthiness interventions.

The sensitivity of memorization scores undermines their reliability in real-world settings. For example, privacy mechanisms relying on memorization thresholds may fail inconsistently across architectures, while interventions targeting “memorized” atypical samples could misidentify noise. Future work should establish standardized, task-agnostic metrics or develop more reliable memorization scores to mitigate these risks.


\subsection{Machine Unlearning}

Machine unlearning, enabling the right to be forgotten, is a critical capability for compliance with regulations like General Data Protection Regulation (GDPR) \citep{mantelero2013eu} and California Consumer Privacy Act (CCPA) \citep{pardau2018california}, which are directly tied to the de-memorization of sensitive information. While existing techniques, such as parameter scrubbing or retraining, aim to remove data traces, their efficacy hinges on understanding how memorization dynamics influence unlearning. For instance, \citet{zhao2024makes} demonstrates that data points with high memorization scores (e.g., atypical or noisy samples) require significantly more effort to unlearn due to their disproportionate influence on model parameters. This asymmetry between the levels of memorization and data resources creates a paradox: models trained to generalize from long-tail data inherently memorize atypical samples, yet precisely, these samples are most resistant to unlearning.

Current approaches mainly treat all data uniformly, ignoring memorization heterogeneity. An open question would be, does unlearning noise (easily memorized but low-risk) demand the same rigor as unlearning sensitive atypical samples (harder to unlearn but high-risk)? Furthermore, the impact unlearning has on fairness remains unexplored: scrubbing minority-class samples to protect privacy may degrade model performance for underrepresented groups. Future work should develop adaptive unlearning strategies that prioritize data based on memorization risk and societal impact, balancing privacy with fairness and robustness.


\section{Conclusion}
In this paper, we investigate the dual role of memorization in trustworthy ML, systematizing its interplay with trustworthiness attributes through a three-level granularity framework of memorization, consisting of class imbalance, atypicality, and noise.  
We have revealed that conflating these levels perpetuates flawed solutions, as memorization being a necessary component for generalization in ML fundamentally conflicts with its risks to privacy, fairness, and robustness. 

From our analysis, two key insights emerge:
1) \textbf{Memorization is not a uniform phenomenon.} Its impact on ML and trustworthiness varies with the level of granularity. 
Leveraging memorization for trustworthiness, therefore,  requires disentangling its effects, i.e., suppressing noise memorization while preserving atypicality memorization, to balance fairness and accuracy.
2) \textbf{Existing theoretical analyses overlook the finer-grained dimension of memorization}, resulting in misaligned performance outcomes. Conducted under the current memorization paradigm, they fail to account for these more nuanced distinctions.

Refining current theoretical frameworks in memorization is essential for research in trustworthy ML. 
Such endeavours also call for introspection and reform within the current body of research, with a focus on re-examining data memorization assumptions, adjusting ML models  to capture the granular nature of memorization, and ensuring that performance metrics align with real-world complexities. 
By adopting this refined paradigm, future research can produce more accurate and robust outcomes.
Specifically, we advocate for adopting granular metrics, such as counterfactual memorization scores that isolate noise effects, and developing theoretical models that formalize memorization’s role in trustworthiness attributes. 
In summary, this survey on current works intersected in memorization and trustworthy ML, affords a new paradigm where models balance accuracy with trustworthiness in high-stakes domains.




\newpage

\bibliographystyle{plainnat}  
\bibliography{dualpath} 

\begin{thebibliography}{222}
\providecommand{\natexlab}[1]{#1}
\providecommand{\url}[1]{\texttt{#1}}
\expandafter\ifx\csname urlstyle\endcsname\relax
  \providecommand{\doi}[1]{doi: #1}\else
  \providecommand{\doi}{doi: \begingroup \urlstyle{rm}\Url}\fi

\bibitem[Aalmoes et~al.(2025)Aalmoes, Duddu, and Boutet]{aalmoes2025alignment}
Jan Aalmoes, Vasisht Duddu, and Antoine Boutet.
\newblock On the alignment of group fairness with attribute privacy.
\newblock In \emph{International Conference on Web Information Systems
  Engineering}, pages 333--348. Springer, 2025.

\bibitem[Abadi et~al.(2016)Abadi, Chu, Goodfellow, McMahan, Mironov, Talwar,
  and Zhang]{abadi2016deep}
Martin Abadi, Andy Chu, Ian Goodfellow, H~Brendan McMahan, Ilya Mironov, Kunal
  Talwar, and Li~Zhang.
\newblock Deep learning with differential privacy.
\newblock In \emph{ACM SIGSAC Conference on Computer and Communications
  Security (CCS)}, pages 308--318, 2016.

\bibitem[Achiam et~al.(2023)Achiam, Adler, Agarwal, Ahmad, Akkaya, Aleman,
  Almeida, Altenschmidt, Altman, Anadkat, et~al.]{achiam2023gpt}
Josh Achiam, Steven Adler, Sandhini Agarwal, Lama Ahmad, Ilge Akkaya,
  Florencia~Leoni Aleman, Diogo Almeida, Janko Altenschmidt, Sam Altman,
  Shyamal Anadkat, et~al.
\newblock Gpt-4 technical report.
\newblock \emph{arXiv preprint arXiv:2303.08774}, 2023.

\bibitem[Agarwal et~al.(2018)Agarwal, Beygelzimer, Dud{\'\i}k, Langford, and
  Wallach]{agarwal2018reduction}
Alekh Agarwal, Alina Beygelzimer, Miroslav Dud{\'\i}k, John Langford, and Hanna
  Wallach.
\newblock A reductions approach to fair classification.
\newblock In \emph{Int. Conf. Mach. Learn. (ICML)}, pages 60--69, 2018.

\bibitem[Agarwal et~al.(2022)Agarwal, D'souza, and
  Hooker]{agarwal2022estimating}
Chirag Agarwal, Daniel D'souza, and Sara Hooker.
\newblock Estimating example difficulty using variance of gradients.
\newblock In \emph{IEEE Conf. Comput. Vis. Pattern Recog. (CVPR)}, pages
  10368--10378, 2022.

\bibitem[Algan and Ulusoy(2021)]{algan2021image}
G{\"o}rkem Algan and Ilkay Ulusoy.
\newblock Image classification with deep learning in the presence of noisy
  labels: A survey.
\newblock \emph{Knowledge-Based Systems}, 215:\penalty0 106771, 2021.

\bibitem[Anderson(2004)]{anderson2004longtail}
Chris Anderson.
\newblock The long tail.
\newblock \emph{Wired}, October 2004.
\newblock URL \url{https://www.wired.com/2004/10/tail/}.

\bibitem[Arpit et~al.(2017)Arpit, Jastrz{\k{e}}bski, Ballas, Krueger, Bengio,
  Kanwal, Maharaj, Fischer, Courville, Bengio, et~al.]{arpit2017closer}
Devansh Arpit, Stanis{\l}aw Jastrz{\k{e}}bski, Nicolas Ballas, David Krueger,
  Emmanuel Bengio, Maxinder~S Kanwal, Tegan Maharaj, Asja Fischer, Aaron
  Courville, Yoshua Bengio, et~al.
\newblock A closer look at memorization in deep networks.
\newblock In \emph{Int. Conf. Mach. Learn. (ICML)}, pages 233--242, 2017.

\bibitem[Babbar and Sch{\"o}lkopf(2019)]{babbar2019data}
Rohit Babbar and Bernhard Sch{\"o}lkopf.
\newblock Data scarcity, robustness and extreme multi-label classification.
\newblock \emph{Machine Learning}, 108\penalty0 (8):\penalty0 1329--1351, 2019.

\bibitem[Bagdasaryan et~al.(2019)Bagdasaryan, Poursaeed, and
  Shmatikov]{bagdasaryan2019differential}
Eugene Bagdasaryan, Omid Poursaeed, and Vitaly Shmatikov.
\newblock Differential privacy has disparate impact on model accuracy.
\newblock \emph{Adv. Neural Inf. Process. Syst. (NeurIPS)}, 32, 2019.

\bibitem[Bai et~al.(2022)Bai, Hsieh, Kan, and Lin]{bai2022reducing}
Andrew Bai, Cho-Jui Hsieh, Wendy Kan, and Hsuan-Tien Lin.
\newblock Reducing training sample memorization in gans by training with
  memorization rejection.
\newblock \emph{arXiv preprint arXiv:2210.12231}, 2022.

\bibitem[Bai et~al.(2024)Bai, Hu, Ye, Li, Wang, and Xu]{bai2024membership}
Li~Bai, Haibo Hu, Qingqing Ye, Haoyang Li, Leixia Wang, and Jianliang Xu.
\newblock Membership inference attacks and defenses in federated learning: A
  survey.
\newblock \emph{Computing Surveys}, 57\penalty0 (4):\penalty0 1--35, 2024.

\bibitem[Baldock et~al.(2021)Baldock, Maennel, and Neyshabur]{baldock2021deep}
Robert Baldock, Hartmut Maennel, and Behnam Neyshabur.
\newblock Deep learning through the lens of example difficulty.
\newblock \emph{Adv. Neural Inf. Process. Syst. (NeurIPS)}, 34:\penalty0
  10876--10889, 2021.

\bibitem[Balle et~al.(2022)Balle, Cherubin, and Hayes]{balle2022reconstructing}
Borja Balle, Giovanni Cherubin, and Jamie Hayes.
\newblock Reconstructing training data with informed adversaries.
\newblock In \emph{IEEE Symposium on Security and Privacy (S\&P)}, pages
  1138--1156, 2022.

\bibitem[Bartlett et~al.(2020)Bartlett, Long, Lugosi, and
  Tsigler]{bartlett2020benign}
Peter~L Bartlett, Philip~M Long, G{\'a}bor Lugosi, and Alexander Tsigler.
\newblock Benign overfitting in linear regression.
\newblock \emph{Proceedings of the National Academy of Sciences}, 117\penalty0
  (48):\penalty0 30063--30070, 2020.

\bibitem[Belkin et~al.(2018)Belkin, Hsu, and Mitra]{belkin2018overfitting}
Mikhail Belkin, Daniel~J Hsu, and Partha Mitra.
\newblock Overfitting or perfect fitting? risk bounds for classification and
  regression rules that interpolate.
\newblock \emph{Adv. Neural Inf. Process. Syst. (NeurIPS)}, 31, 2018.

\bibitem[Belkin et~al.(2019)Belkin, Hsu, Ma, and Mandal]{belkin2019reconciling}
Mikhail Belkin, Daniel Hsu, Siyuan Ma, and Soumik Mandal.
\newblock Reconciling modern machine-learning practice and the classical
  bias--variance trade-off.
\newblock \emph{Proceedings of the National Academy of Sciences}, 116\penalty0
  (32):\penalty0 15849--15854, 2019.

\bibitem[Benz et~al.(2021)Benz, Zhang, Karjauv, and Kweon]{benz2021robustness}
Philipp Benz, Chaoning Zhang, Adil Karjauv, and In~So Kweon.
\newblock Robustness may be at odds with fairness: An empirical study on
  class-wise accuracy.
\newblock In \emph{NeurIPS Workshop on pre-registration in machine learning},
  pages 325--342. PMLR, 2021.

\bibitem[Bird(2024)]{bird-2024-must}
Steven Bird.
\newblock Must {NLP} be extractive?
\newblock In Lun-Wei Ku, Andre Martins, and Vivek Srikumar, editors,
  \emph{Proceedings of the 62nd Annual Meeting of the Association for
  Computational Linguistics (Volume 1: Long Papers)}, pages 14915--14929, 2024.

\bibitem[Blevins et~al.(2024)Blevins, Limisiewicz, Gururangan, Li, Gonen,
  Smith, and Zettlemoyer]{blevins-etal-2024-breaking}
Terra Blevins, Tomasz Limisiewicz, Suchin Gururangan, Margaret Li, Hila Gonen,
  Noah~A. Smith, and Luke Zettlemoyer.
\newblock Breaking the curse of multilinguality with cross-lingual expert
  language models.
\newblock In Yaser Al-Onaizan, Mohit Bansal, and Yun-Nung Chen, editors,
  \emph{Proceedings of the 2024 Conference on Empirical Methods in Natural
  Language Processing}, pages 10822--10837, 2024.

\bibitem[Boenisch et~al.(2021)Boenisch, Sperl, and
  B{\"o}ttinger]{boenisch2021gradient}
Franziska Boenisch, Philip Sperl, and Konstantin B{\"o}ttinger.
\newblock Gradient masking and the underestimated robustness threats of
  differential privacy in deep learning.
\newblock \emph{arXiv preprint arXiv:2105.07985}, 2021.

\bibitem[Brendel et~al.(2017)Brendel, Rauber, and Bethge]{brendel2017decision}
Wieland Brendel, Jonas Rauber, and Matthias Bethge.
\newblock Decision-based adversarial attacks: Reliable attacks against
  black-box machine learning models.
\newblock \emph{arXiv preprint arXiv:1712.04248}, 2017.

\bibitem[Brown et~al.(2021)Brown, Bun, Feldman, Smith, and
  Talwar]{brown2021memorization}
Gavin Brown, Mark Bun, Vitaly Feldman, Adam Smith, and Kunal Talwar.
\newblock When is memorization of irrelevant training data necessary for
  high-accuracy learning?
\newblock In \emph{SIGACT Symposium on Theory of Computing}, pages 123--132.
  ACM, 2021.

\bibitem[Buda et~al.(2018)Buda, Maki, and Mazurowski]{buda2018systematic}
Mateusz Buda, Atsuto Maki, and Maciej~A Mazurowski.
\newblock A systematic study of the class imbalance problem in convolutional
  neural networks.
\newblock \emph{Neural networks}, 106:\penalty0 249--259, 2018.

\bibitem[{C. Dwork}(2006)]{dwork2006}
{C. Dwork}.
\newblock Differential privacy.
\newblock In \emph{ICALP, pp. 1--12}, 2006.
\newblock ISBN 978-3-540-35908-1.

\bibitem[{C. Dwork, and G.N. Rothblum}(2016)]{dwork2016concentrated}
{C. Dwork, and G.N. Rothblum}.
\newblock Concentrated differential privacy.
\newblock \emph{arXiv preprint arXiv:1603.01887}, 2016.

\bibitem[{C. Dwork and J. Lei}(2009)]{dwork2009differential}
{C. Dwork and J. Lei}.
\newblock Differential privacy and robust statistics.
\newblock In \emph{Proc. 41st Annu. ACM Symp. Theory Comput.}, pages 371--380,
  2009.

\bibitem[{C. Dwork, F. McSherry, K. Nissim, A.
  Smith}(2006)]{dwork2006calibrating}
{C. Dwork, F. McSherry, K. Nissim, A. Smith}.
\newblock Calibrating noise to sensitivity in private data analysis.
\newblock In \emph{Proc. Theory of Cryptography Conf. , pp. 265-284}, 2006.

\bibitem[Carlini and Wagner(2017)]{carlini2017towards}
Nicholas Carlini and David Wagner.
\newblock Towards evaluating the robustness of neural networks.
\newblock In \emph{IEEE Symposium on Security and Privacy (S\&P)}, pages
  39--57, 2017.

\bibitem[Carlini et~al.(2019)Carlini, Erlingsson, and
  Papernot]{carlini2019distribution}
Nicholas Carlini, Ulfar Erlingsson, and Nicolas Papernot.
\newblock Distribution density, tails, and outliers in machine learning:
  Metrics and applications.
\newblock \emph{arXiv preprint arXiv:1910.13427}, 2019.

\bibitem[Carlini et~al.(2021)Carlini, Tramer, Wallace, Jagielski, Herbert-Voss,
  Lee, Roberts, Brown, Song, Erlingsson, et~al.]{carlini2021extracting}
Nicholas Carlini, Florian Tramer, Eric Wallace, Matthew Jagielski, Ariel
  Herbert-Voss, Katherine Lee, Adam Roberts, Tom Brown, Dawn Song, Ulfar
  Erlingsson, et~al.
\newblock Extracting training data from large language models.
\newblock In \emph{USENIX Security Symposium}, pages 2633--2650, 2021.

\bibitem[Carlini et~al.(2022{\natexlab{a}})Carlini, Chien, Nasr, Song, Terzis,
  and Tramer]{carlini2022membership}
Nicholas Carlini, Steve Chien, Milad Nasr, Shuang Song, Andreas Terzis, and
  Florian Tramer.
\newblock Membership inference attacks from first principles.
\newblock In \emph{IEEE Symposium on Security and Privacy (S\&P)}, pages
  1897--1914, 2022{\natexlab{a}}.

\bibitem[Carlini et~al.(2022{\natexlab{b}})Carlini, Jagielski, Zhang, Papernot,
  Terzis, and Tramer]{carlini2022privacy}
Nicholas Carlini, Matthew Jagielski, Chiyuan Zhang, Nicolas Papernot, Andreas
  Terzis, and Florian Tramer.
\newblock The privacy onion effect: Memorization is relative.
\newblock \emph{Adv. Neural Inf. Process. Syst. (NeurIPS)}, 35:\penalty0
  13263--13276, 2022{\natexlab{b}}.

\bibitem[Carlini et~al.(2023)Carlini, Hayes, Nasr, Jagielski, Sehwag, Tramer,
  Balle, Ippolito, and Wallace]{carlini2023extracting}
Nicolas Carlini, Jamie Hayes, Milad Nasr, Matthew Jagielski, Vikash Sehwag,
  Florian Tramer, Borja Balle, Daphne Ippolito, and Eric Wallace.
\newblock Extracting training data from diffusion models.
\newblock In \emph{USENIX Security Symposium}, pages 5253--5270, 2023.

\bibitem[Caton and Haas(2024)]{caton2024fairness}
Simon Caton and Christian Haas.
\newblock Fairness in machine learning: A survey.
\newblock \emph{Computing Surveys}, 56\penalty0 (7):\penalty0 1--38, 2024.

\bibitem[Cavalin et~al.(2024)Cavalin, Domingues, Pinhanez, and
  Nogima]{cavalin2024fixing}
Paulo Cavalin, Pedro~Henrique Domingues, Claudio Pinhanez, and Julio Nogima.
\newblock Fixing rogue memorization in many-to-one multilingual translators of
  extremely-low-resource languages by rephrasing training samples.
\newblock In \emph{Proceedings of the 2024 Conference of the North American
  Chapter of the Association for Computational Linguistics: Human Language
  Technologies (Volume 1: Long Papers)}, pages 4503--4514, 2024.

\bibitem[Chang and Shokri(2021)]{chang2021privacy}
Hongyan Chang and Reza Shokri.
\newblock On the privacy risks of algorithmic fairness.
\newblock In \emph{EuroS\&P}, pages 292--303. IEEE, 2021.

\bibitem[Chang et~al.(2020)Chang, Nguyen, Murakonda, Kazemi, and
  Shokri]{chang2020adversarial}
Hongyan Chang, Ta~Duy Nguyen, Sasi~Kumar Murakonda, Ehsan Kazemi, and Reza
  Shokri.
\newblock On adversarial bias and the robustness of fair machine learning.
\newblock \emph{arXiv preprint arXiv:2006.08669}, 2020.

\bibitem[Chang et~al.(2023)Chang, Arnett, Tu, and
  Bergen]{chang2023multilingualitycurselanguagemodeling}
Tyler~A. Chang, Catherine Arnett, Zhuowen Tu, and Benjamin~K. Bergen.
\newblock When is multilinguality a curse? language modeling for 250 high- and
  low-resource languages.
\newblock \emph{arXiv preprint arXiv:2311.09205}, 2023.

\bibitem[Chataigner et~al.(2024)Chataigner, Ta{\"\i}k, and
  Farnadi]{chataigner2024multilingual}
Cl{\'e}a Chataigner, Afaf Ta{\"\i}k, and Golnoosh Farnadi.
\newblock Multilingual hallucination gaps in large language models.
\newblock \emph{arXiv preprint arXiv:2410.18270}, 2024.

\bibitem[Chatterjee and Zielinski(2020)]{chatterjee2020making}
Satrajit Chatterjee and Piotr Zielinski.
\newblock Making coherence out of nothing at all: measuring the evolution of
  gradient alignment.
\newblock \emph{arXiv preprint arXiv:2008.01217}, 2020.

\bibitem[Chatterjee and Zielinski(2022)]{chatterjee2022generalization}
Satrajit Chatterjee and Piotr Zielinski.
\newblock On the generalization mystery in deep learning.
\newblock \emph{arXiv preprint arXiv:2203.10036}, 2022.

\bibitem[Chen et~al.(2024{\natexlab{a}})Chen, Biswas, Lent, and
  Bjerva]{chen2024against}
Yiyi Chen, Russa Biswas, Heather Lent, and Johannes Bjerva.
\newblock Against all odds: Overcoming typology, script, and language confusion
  in multilingual embedding inversion attacks.
\newblock \emph{arXiv preprint arXiv:2408.11749}, 2024{\natexlab{a}}.

\bibitem[Chen et~al.(2024{\natexlab{b}})Chen, Lent, and Bjerva]{chen2024text}
Yiyi Chen, Heather Lent, and Johannes Bjerva.
\newblock Text embedding inversion security for multilingual language models.
\newblock In \emph{Proceedings of the 62nd Annual Meeting of the Association
  for Computational Linguistics (Volume 1: Long Papers)}, pages 7808--7827,
  2024{\natexlab{b}}.

\bibitem[Chen et~al.(2025)Chen, Li, Biswas, and
  Bjerva]{chen2025largelanguagemodelseasily}
Yiyi Chen, Qiongxiu Li, Russa Biswas, and Johannes Bjerva.
\newblock Large language models are easily confused: A quantitative metric,
  security implications and typological analysis.
\newblock In \emph{Findings of the Association for Computational Linguistics:
  NAACL, to appear}, 2025.

\bibitem[Choi et~al.(2021)Choi, Kim, Joe, Min, and
  Gwon]{choi2021analyzingzeroshotcrosslingualtransfer}
Hyunjin Choi, Judong Kim, Seongho Joe, Seungjai Min, and Youngjune Gwon.
\newblock Analyzing zero-shot cross-lingual transfer in supervised nlp tasks.
\newblock \emph{arXiv preprint arXiv:2101.10649}, 2021.

\bibitem[Chuang and Mroueh(2021)]{chuang2021fair}
Ching-Yao Chuang and Youssef Mroueh.
\newblock Fair mixup: Fairness via interpolation.
\newblock \emph{arXiv preprint arXiv:2103.06503}, 2021.

\bibitem[Cohen et~al.(2019)Cohen, Rosenfeld, and Kolter]{cohen2019certified}
Jeremy Cohen, Elan Rosenfeld, and Zico Kolter.
\newblock Certified adversarial robustness via randomized smoothing.
\newblock In \emph{Int. Conf. Mach. Learn. (ICML)}, pages 1310--1320, 2019.

\bibitem[Conneau et~al.(2018)Conneau, Rinott, Lample, Williams, Bowman,
  Schwenk, and Stoyanov]{conneau-etal-2018-xnli}
Alexis Conneau, Ruty Rinott, Guillaume Lample, Adina Williams, Samuel Bowman,
  Holger Schwenk, and Veselin Stoyanov.
\newblock {XNLI}: Evaluating cross-lingual sentence representations.
\newblock In Ellen Riloff, David Chiang, Julia Hockenmaier, and Jun{'}ichi
  Tsujii, editors, \emph{Proceedings of the 2018 Conference on Empirical
  Methods in Natural Language Processing}, pages 2475--2485, 2018.

\bibitem[Conneau et~al.(2020)Conneau, Khandelwal, Goyal, Chaudhary, Wenzek,
  Guzm{\'a}n, Grave, Ott, Zettlemoyer, and
  Stoyanov]{conneau-etal-2020-unsupervised}
Alexis Conneau, Kartikay Khandelwal, Naman Goyal, Vishrav Chaudhary, Guillaume
  Wenzek, Francisco Guzm{\'a}n, Edouard Grave, Myle Ott, Luke Zettlemoyer, and
  Veselin Stoyanov.
\newblock Unsupervised cross-lingual representation learning at scale.
\newblock In Dan Jurafsky, Joyce Chai, Natalie Schluter, and Joel Tetreault,
  editors, \emph{Proceedings of the 58th Annual Meeting of the Association for
  Computational Linguistics}, pages 8440--8451, 2020.

\bibitem[Cook and Weisberg(1980)]{cook1980characterizations}
R~Dennis Cook and Sanford Weisberg.
\newblock Characterizations of an empirical influence function for detecting
  influential cases in regression.
\newblock \emph{Technometrics}, 22\penalty0 (4):\penalty0 495--508, 1980.

\bibitem[Cui et~al.(2019)Cui, Jia, Lin, Song, and Belongie]{cui2019class}
Yin Cui, Menglin Jia, Tsung-Yi Lin, Yang Song, and Serge Belongie.
\newblock Class-balanced loss based on effective number of samples.
\newblock In \emph{IEEE Conf. Comput. Vis. Pattern Recog. (CVPR)}, pages
  9268--9277, 2019.

\bibitem[Cummings et~al.(2019)Cummings, Gupta, Kimpara, and
  Morgenstern]{cummings2019compatibility}
Rachel Cummings, Varun Gupta, Dhamma Kimpara, and Jamie Morgenstern.
\newblock On the compatibility of privacy and fairness.
\newblock In \emph{Adjunct publication of the 27th conference on user modeling,
  adaptation and personalization}, pages 309--315, 2019.

\bibitem[Dankers and Titov(2024)]{dankers2024generalisation}
Verna Dankers and Ivan Titov.
\newblock Generalisation first, memorisation second? memorisation localisation
  for natural language classification tasks.
\newblock \emph{arXiv preprint arXiv:2408.04965}, 2024.

\bibitem[Deng et~al.(2023)Deng, Zhang, Pan, and Bing]{deng2023multilingual}
Yue Deng, Wenxuan Zhang, Sinno~Jialin Pan, and Lidong Bing.
\newblock Multilingual jailbreak challenges in large language models.
\newblock \emph{arXiv preprint arXiv:2310.06474}, 2023.

\bibitem[Dong et~al.(2021)Dong, Xu, Yang, Pang, Deng, Su, and
  Zhu]{dong2021exploring}
Yinpeng Dong, Ke~Xu, Xiao Yang, Tianyu Pang, Zhijie Deng, Hang Su, and Jun Zhu.
\newblock Exploring memorization in adversarial training.
\newblock \emph{arXiv preprint arXiv:2106.01606}, 2021.

\bibitem[Doshi-Velez and Kim(2017)]{doshi2017towards}
Finale Doshi-Velez and Been Kim.
\newblock Towards a rigorous science of interpretable machine learning.
\newblock \emph{arXiv preprint arXiv:1702.08608}, 2017.

\bibitem[D'souza et~al.(2021)D'souza, Nussbaum, Agarwal, and Hooker]{d2021tale}
Daniel D'souza, Zach Nussbaum, Chirag Agarwal, and Sara Hooker.
\newblock A tale of two long tails.
\newblock \emph{arXiv preprint arXiv:2107.13098}, 2021.

\bibitem[Dwork et~al.(2012)Dwork, Hardt, Pitassi, Reingold, and
  Zemel]{dwork2012fairness}
Cynthia Dwork, Moritz Hardt, Toniann Pitassi, Omer Reingold, and Richard Zemel.
\newblock Fairness through awareness.
\newblock In \emph{Proceedings of the 3rd innovations in theoretical computer
  science conference}, pages 214--226, 2012.

\bibitem[Dziugaite(2020)]{dziugaite2020revisiting}
Gintare~Karolina Dziugaite.
\newblock \emph{Revisiting generalization for deep learning: PAC-Bayes, flat
  minima, and generative models}.
\newblock PhD thesis, 2020.

\bibitem[Eykholt et~al.(2018)Eykholt, Evtimov, Fernandes, Li, Rahmati, Xiao,
  Prakash, Kohno, and Song]{eykholt2018robust}
Kevin Eykholt, Ivan Evtimov, Earlence Fernandes, Bo~Li, Amir Rahmati, Chaowei
  Xiao, Atul Prakash, Tadayoshi Kohno, and Dawn Song.
\newblock Robust physical-world attacks on deep learning models.
\newblock \emph{IEEE Conf. Comput. Vis. Pattern Recog. (CVPR)}, pages
  1625--1634, 2018.

\bibitem[Farrand et~al.(2020)Farrand, Mireshghallah, Singh, and
  Trask]{farrand2020neither}
Tom Farrand, Fatemehsadat Mireshghallah, Sahib Singh, and Andrew Trask.
\newblock Neither private nor fair: Impact of data imbalance on utility and
  fairness in differential privacy.
\newblock In \emph{Proceedings of the 2020 workshop on privacy-preserving
  machine learning in practice}, pages 15--19, 2020.

\bibitem[Feldman(2020)]{feldman2020does}
Vitaly Feldman.
\newblock Does learning require memorization? a short tale about a long tail.
\newblock In \emph{SIGACT Symposium on Theory of Computing}, pages 954--959.
  ACM, 2020.

\bibitem[Feldman and Zhang(2020)]{feldman2020neural}
Vitaly Feldman and Chiyuan Zhang.
\newblock What neural networks memorize and why: Discovering the long tail via
  influence estimation.
\newblock \emph{Adv. Neural Inf. Process. Syst. (NeurIPS)}, 33:\penalty0
  2881--2891, 2020.

\bibitem[Feng et~al.(2021)Feng, Guo, Benitez-Quiroz, and
  Martinez]{feng2021gans}
Qianli Feng, Chenqi Guo, Fabian Benitez-Quiroz, and Aleix~M Martinez.
\newblock When do gans replicate? on the choice of dataset size.
\newblock In \emph{Int. Conf. Comput. Vis. (ICCV)}, pages 6701--6710, 2021.

\bibitem[Feng et~al.(2023)Feng, Hebbar, Mehlman, Shi, Kommineni, Narayanan,
  et~al.]{feng2023review}
Tiantian Feng, Rajat Hebbar, Nicholas Mehlman, Xuan Shi, Aditya Kommineni,
  Shrikanth Narayanan, et~al.
\newblock A review of speech-centric trustworthy machine learning: Privacy,
  safety, and fairness.
\newblock \emph{APSIPA Transactions on Signal and Information Processing},
  12\penalty0 (3), 2023.

\bibitem[Floridi and Chiriatti(2020)]{floridi2020gpt}
Luciano Floridi and Massimo Chiriatti.
\newblock Gpt-3: Its nature, scope, limits, and consequences.
\newblock \emph{Minds and Machines}, 30:\penalty0 681--694, 2020.

\bibitem[Fredrikson et~al.(2015)Fredrikson, Jha, and
  Ristenpart]{fredrikson2015model}
Matt Fredrikson, Somesh Jha, and Thomas Ristenpart.
\newblock Model inversion attacks that exploit confidence information and basic
  countermeasures.
\newblock In \emph{ACM SIGSAC Conference on Computer and Communications
  Security (CCS)}, pages 1322--1333, 2015.

\bibitem[Fu et~al.(2022)Fu, He, Liu, Shen, and Tao]{fu2022robust}
Shaopeng Fu, Fengxiang He, Yang Liu, Li~Shen, and Dacheng Tao.
\newblock Robust unlearnable examples: Protecting data against adversarial
  learning.
\newblock \emph{arXiv preprint arXiv:2203.14533}, 2022.

\bibitem[Gao et~al.(2023)Gao, Li, Zhu, Wu, Jiang, and Xia]{gao2023not}
Yinghua Gao, Yiming Li, Linghui Zhu, Dongxian Wu, Yong Jiang, and Shu-Tao Xia.
\newblock Not all samples are born equal: Towards effective clean-label
  backdoor attacks.
\newblock \emph{Pattern Recognition}, 139:\penalty0 109512, 2023.

\bibitem[Garg et~al.(2023)Garg, Ravikumar, and Roy]{garg2023memorization}
Isha Garg, Deepak Ravikumar, and Kaushik Roy.
\newblock Memorization through the lens of curvature of loss function around
  samples.
\newblock \emph{arXiv preprint arXiv:2307.05831}, 2023.

\bibitem[Goodfellow et~al.(2015)Goodfellow, Shlens, and
  Szegedy]{goodfellow2014explaining}
Ian~J Goodfellow, Jonathon Shlens, and Christian Szegedy.
\newblock Explaining and harnessing adversarial examples.
\newblock In \emph{Int. Conf. Learn. Represent. (ICLR)}, 2015.

\bibitem[Guerreiro et~al.(2023)Guerreiro, Alves, Waldendorf, Haddow, Birch,
  Colombo, and Martins]{guerreiro2023hallucinations}
Nuno~M Guerreiro, Duarte~M Alves, Jonas Waldendorf, Barry Haddow, Alexandra
  Birch, Pierre Colombo, and Andr{\'e}~FT Martins.
\newblock Hallucinations in large multilingual translation models.
\newblock \emph{Transactions of the Association for Computational Linguistics},
  11:\penalty0 1500--1517, 2023.

\bibitem[Han et~al.(2018)Han, Yao, Yu, Niu, Xu, Hu, Tsang, and
  Sugiyama]{han2018co}
Bo~Han, Quanming Yao, Xingrui Yu, Gang Niu, Miao Xu, Weihua Hu, Ivor Tsang, and
  Masashi Sugiyama.
\newblock Co-teaching: Robust training of deep neural networks with extremely
  noisy labels.
\newblock \emph{Adv. Neural Inf. Process. Syst. (NeurIPS)}, 31, 2018.

\bibitem[Hardt et~al.(2016)Hardt, Price, and Srebro]{hardt2016equality}
Moritz Hardt, Eric Price, and Nathan Srebro.
\newblock Equality of opportunity in supervised learning.
\newblock In \emph{Adv. Neural Inf. Process. Syst. (NeurIPS)}, pages
  3315--3323, 2016.

\bibitem[He and Garcia(2009)]{he2009learning}
Haibo He and Edwardo~A Garcia.
\newblock Learning from imbalanced data.
\newblock \emph{TKDE}, 21\penalty0 (9):\penalty0 1263--1284, 2009.

\bibitem[He et~al.(2024)He, Wang, Xu, Minervini, Stenetorp, Rubinstein, and
  Cohn]{he2024transferring}
Xuanli He, Jun Wang, Qiongkai Xu, Pasquale Minervini, Pontus Stenetorp,
  Benjamin~IP Rubinstein, and Trevor Cohn.
\newblock Transferring troubles: Cross-lingual transferability of backdoor
  attacks in llms with instruction tuning.
\newblock \emph{arXiv preprint arXiv:2404.19597}, 2024.

\bibitem[He et~al.(2019)He, Zhang, and Lee]{he2019model}
Z.~He, T.~Zhang, and R.~Lee.
\newblock Model inversion attacks against collaborative inference.
\newblock In \emph{35th Annu. Comput. Secur. Appl. Conf.}, pages 148--162,
  2019.

\bibitem[Hintersdorf et~al.(2022)Hintersdorf, Struppek, and
  Kersting]{hintersdorftrust}
Dominik Hintersdorf, Lukas Struppek, and Kristian Kersting.
\newblock To trust or not to trust prediction scores for membership inference
  attacks.
\newblock In \emph{Int. Joint Conf. Artif. Intell. (IJCAI)}, pages 3043--3049,
  2022.

\bibitem[Hong et~al.(2020)Hong, Chandrasekaran, Kaya, Dumitra{\c{s}}, and
  Papernot]{hong2020effectiveness}
Sanghyun Hong, Varun Chandrasekaran, Yi{\u{g}}itcan Kaya, Tudor Dumitra{\c{s}},
  and Nicolas Papernot.
\newblock On the effectiveness of mitigating data poisoning attacks with
  gradient shaping.
\newblock \emph{arXiv preprint arXiv:2002.11497}, 2020.

\bibitem[Hu et~al.(2021)Hu, Salcic, Dobbie, and Zhang]{hu2021membership}
Hongsheng Hu, Zoran Salcic, Gillian Dobbie, and Xuyun Zhang.
\newblock Membership inference attacks on machine learning: A survey.
\newblock \emph{Computing Surveys}, 2021.

\bibitem[{I. Damg{\aa}rd, V. Pastro, N. Smart, and S.
  Zakarias}(2012)]{damgaard2012multiparty}
{I. Damg{\aa}rd, V. Pastro, N. Smart, and S. Zakarias}.
\newblock Multiparty computation from somewhat homomorphic encryption.
\newblock In \emph{Advances in Cryptology--CRYPTO, pp. 643--662}. Springer,
  2012.

\bibitem[Islam et~al.(2025)Islam, Lauscher, and
  Glavaš]{islam2025llmshallucinatelanguagesmultilingual}
Saad Obaid~ul Islam, Anne Lauscher, and Goran Glavaš.
\newblock How much do llms hallucinate across languages? on multilingual
  estimation of llm hallucination in the wild.
\newblock \emph{arXiv preprint arXiv:2502.12769}, 2025.

\bibitem[Jagielski et~al.(2018)Jagielski, Oprea, Biggio, Liu, Nita-Rotaru, and
  Li]{jagielski2018manipulating}
Matthew Jagielski, Alina Oprea, Battista Biggio, Chang Liu, Cristina
  Nita-Rotaru, and Bo~Li.
\newblock Manipulating machine learning: Poisoning attacks and countermeasures
  for regression learning.
\newblock In \emph{IEEE Symposium on Security and Privacy (S\&P)}, pages
  19--35, 2018.

\bibitem[Jalalzai et~al.(2020)Jalalzai, Colombo, Clavel, Gaussier, Varni,
  Vignon, and Sabourin]{jalalzai2020heavy}
Hamid Jalalzai, Pierre Colombo, Chlo{\'e} Clavel, Eric Gaussier, Giovanna
  Varni, Emmanuel Vignon, and Anne Sabourin.
\newblock Heavy-tailed representations, text polarity classification \& data
  augmentation.
\newblock \emph{Adv. Neural Inf. Process. Syst. (NeurIPS)}, 33:\penalty0
  4295--4307, 2020.

\bibitem[Jia et~al.(2019)Jia, Salem, Backes, Zhang, and Gong]{jia2019memguard}
Jinyuan Jia, Ahmed Salem, Michael Backes, Yang Zhang, and Neil~Zhenqiang Gong.
\newblock Memguard: Defending against black-box membership inference attacks
  via adversarial examples.
\newblock In \emph{ACM SIGSAC Conference on Computer and Communications
  Security (CCS)}, pages 259--274, 2019.

\bibitem[Jiang et~al.(2020)Jiang, Zhang, Talwar, and Mozer]{jiang2020exploring}
Ziheng Jiang, Chiyuan Zhang, Kunal Talwar, and Michael~C Mozer.
\newblock Exploring the memorization-generalization continuum in deep learning.
\newblock \emph{arXiv preprint arXiv:2002.03206}, 2020.

\bibitem[Joshi et~al.(2020)Joshi, Santy, Budhiraja, Bali, and
  Choudhury]{joshi-etal-2020-state}
Pratik Joshi, Sebastin Santy, Amar Budhiraja, Kalika Bali, and Monojit
  Choudhury.
\newblock The state and fate of linguistic diversity and inclusion in the {NLP}
  world.
\newblock In Dan Jurafsky, Joyce Chai, Natalie Schluter, and Joel Tetreault,
  editors, \emph{Proceedings of the 58th Annual Meeting of the Association for
  Computational Linguistics}, pages 6282--6293, 2020.

\bibitem[Kamiran and Calders(2012)]{kamiran2012data}
Faisal Kamiran and Toon Calders.
\newblock Data preprocessing techniques for classification without
  discrimination.
\newblock \emph{Knowledge and information systems}, 33\penalty0 (1):\penalty0
  1--33, 2012.

\bibitem[Kang et~al.(2019)Kang, Xie, Rohrbach, Yan, Gordo, Feng, and
  Kalantidis]{kang2019decoupling}
Bingyi Kang, Saining Xie, Marcus Rohrbach, Zhicheng Yan, Albert Gordo, Jiashi
  Feng, and Yannis Kalantidis.
\newblock Decoupling representation and classifier for long-tailed recognition.
\newblock \emph{arXiv preprint arXiv:1910.09217}, 2019.

\bibitem[Kang et~al.(2024)Kang, Blevins, and
  Zettlemoyer]{kang2024comparinghallucinationdetectionmetrics}
Haoqiang Kang, Terra Blevins, and Luke Zettlemoyer.
\newblock Comparing hallucination detection metrics for multilingual
  generation.
\newblock \emph{arXiv preprint arXiv:2402.10496}, 2024.

\bibitem[Kaur et~al.(2022)Kaur, Uslu, Rittichier, and
  Durresi]{kaur2022trustworthy}
Davinder Kaur, Suleyman Uslu, Kaley~J Rittichier, and Arjan Durresi.
\newblock Trustworthy artificial intelligence: a review.
\newblock \emph{Computing Survey}, 55\penalty0 (2):\penalty0 1--38, 2022.

\bibitem[Khalili et~al.(2021)Khalili, Zhang, Abroshan, and
  Sojoudi]{khalili2021improving}
Mohammad~Mahdi Khalili, Xueru Zhang, Mahed Abroshan, and Somayeh Sojoudi.
\newblock Improving fairness and privacy in selection problems.
\newblock In \emph{AAAI Conf. Artif. Intell. (AAAI)}, volume~35, pages
  8092--8100, 2021.

\bibitem[Kishida and Nakayama(2019)]{kishida2019empirical}
Ikki Kishida and Hideki Nakayama.
\newblock Empirical study of easy and hard examples in cnn training.
\newblock In \emph{Neural Information Processing (ICONIP)}, pages 179--188,
  2019.

\bibitem[Koh and Liang(2017)]{koh2017understanding}
Pang~Wei Koh and Percy Liang.
\newblock Understanding black-box predictions via influence functions.
\newblock In \emph{Int. Conf. Mach. Learn. (ICML)}, pages 1885--1894, 2017.

\bibitem[Kolter and Maloof(2006)]{kolter2006learning}
Jeremy~Z. Kolter and Marcus~A. Maloof.
\newblock Learning to detect and classify malicious executables in the wild.
\newblock \emph{Journal of Machine Learning Research}, 7\penalty0 (12), 2006.

\bibitem[Krizhevsky et~al.(2009)Krizhevsky, Hinton,
  et~al.]{krizhevsky2009learning}
Alex Krizhevsky, Geoffrey Hinton, et~al.
\newblock Learning multiple layers of features from tiny images.
\newblock 2009.

\bibitem[Kulynych et~al.(2019)Kulynych, Yaghini, Cherubin, Veale, and
  Troncoso]{kulynych2019disparate}
Bogdan Kulynych, Mohammad Yaghini, Giovanni Cherubin, Michael Veale, and
  Carmela Troncoso.
\newblock Disparate vulnerability to membership inference attacks.
\newblock \emph{arXiv preprint arXiv:1906.00389}, 2019.

\bibitem[Kwok et~al.(2024)Kwok, Anand, Frankle, Dziugaite, and
  Rolnick]{kwok2024dataset}
Devin Kwok, Nikhil Anand, Jonathan Frankle, Gintare~Karolina Dziugaite, and
  David Rolnick.
\newblock Dataset difficulty and the role of inductive bias.
\newblock \emph{arXiv preprint arXiv:2401.01867}, 2024.

\bibitem[L.~Melis and Shmatikov(2019)]{melis2019exploiting}
E.~De~Cristofaro L.~Melis, C.~Song and V.~Shmatikov.
\newblock Exploiting unintended feature leakage in collaborative learning.
\newblock In \emph{SP}, page 691–706. IEEE, 2019.

\bibitem[Lazarevic et~al.(2003)Lazarevic, Ert{"{o}}z, Kumar, Ozgur, and
  Srivastava]{lazarevic2003comparative}
Aleksandar Lazarevic, Levent Ert{"{o}}z, Vipin Kumar, Aysel Ozgur, and Jaideep
  Srivastava.
\newblock A comparative study of anomaly detection schemes in network intrusion
  detection.
\newblock In \emph{Proceedings of the SIAM International Conference on Data
  Mining (SDM)}, pages 25--36, 2003.

\bibitem[Li et~al.(2023{\natexlab{a}})Li, Qi, Liu, Di, Liu, Pei, Yi, and
  Zhou]{li2023trustworthy}
Bo~Li, Peng Qi, Bo~Liu, Shuai Di, Jingen Liu, Jiquan Pei, Jinfeng Yi, and Bowen
  Zhou.
\newblock Trustworthy ai: From principles to practices.
\newblock \emph{Computing Surveys}, 55\penalty0 (9):\penalty0 1--46,
  2023{\natexlab{a}}.

\bibitem[Li et~al.(2023{\natexlab{b}})Li, Xu, and Zhang]{li2023alleviating}
Guanlin Li, Guowen Xu, and Tianwei Zhang.
\newblock Alleviating the effect of data imbalance on adversarial training.
\newblock \emph{arXiv preprint arXiv:2307.10205}, 2023{\natexlab{b}}.

\bibitem[Li et~al.(2023{\natexlab{c}})Li, Xie, and Li]{li2023sok}
Linyi Li, Tao Xie, and Bo~Li.
\newblock Sok: Certified robustness for deep neural networks.
\newblock In \emph{IEEE Symposium on Security and Privacy (S\&P)}, pages
  1289--1310, 2023{\natexlab{c}}.

\bibitem[Li et~al.(2024{\natexlab{a}})Li, Li, Hu, and Hu]{li2024privacy}
Xiao Li, Qiongxiu Li, Zhanhao Hu, and Xiaolin Hu.
\newblock On the privacy effect of data enhancement via the lens of
  memorization.
\newblock \emph{IEEE Trans. Inf. Forensics Secur. (TIFS)}, 2024{\natexlab{a}}.

\bibitem[Li et~al.(2024{\natexlab{b}})Li, Li, Li, Lee, Cui, and
  Hu]{li2024faster}
Xiao Li, Zhuhong Li, Qiongxiu Li, Bingze Lee, Jinghao Cui, and Xiaolin Hu.
\newblock Faster-gcg: Efficient discrete optimization jailbreak attacks against
  aligned large language models.
\newblock \emph{arXiv preprint arXiv:2410.15362}, 2024{\natexlab{b}}.

\bibitem[Li et~al.(2025)Li, Sun, Chen, Li, Liu, He, Shi, and Hu]{li2024adbm}
Xiao Li, Wenxuan Sun, Huanran Chen, Qiongxiu Li, Yining Liu, Yingzhe He, Jie
  Shi, and Xiaolin Hu.
\newblock Adbm: Adversarial diffusion bridge model for reliable adversarial
  purification.
\newblock \emph{Int. Conf. Learn. Represent. (ICLR)}, 2025.

\bibitem[Liu et~al.(2024)Liu, Feng, Xue, Wang, Wu, Lu, Zhao, Deng, Zhang, Ruan,
  et~al.]{liu2024deepseek}
Aixin Liu, Bei Feng, Bing Xue, Bingxuan Wang, Bochao Wu, Chengda Lu, Chenggang
  Zhao, Chengqi Deng, Chenyu Zhang, Chong Ruan, et~al.
\newblock Deepseek-v3 technical report.
\newblock \emph{arXiv preprint arXiv:2412.19437}, 2024.

\bibitem[Liu et~al.(2021)Liu, Lin, and Jaggi]{liu2021understanding}
Futong Liu, Tao Lin, and Martin Jaggi.
\newblock Understanding memorization from the perspective of optimization via
  efficient influence estimation.
\newblock \emph{arXiv preprint arXiv:2112.08798}, 2021.

\bibitem[Liu et~al.(2023)Liu, Chaudhary, and Wang]{liu2023towards}
Haoyang Liu, Maheep Chaudhary, and Haohan Wang.
\newblock Towards trustworthy and aligned machine learning: A data-centric
  survey with causality perspectives.
\newblock \emph{arXiv preprint arXiv:2307.16851}, 2023.

\bibitem[Liu et~al.(2020)Liu, Niles-Weed, Razavian, and
  Fernandez-Granda]{liu2020early}
Sheng Liu, Jonathan Niles-Weed, Narges Razavian, and Carlos Fernandez-Granda.
\newblock Early-learning regularization prevents memorization of noisy labels.
\newblock \emph{Adv. Neural Inf. Process. Syst. (NeurIPS)}, 33:\penalty0
  20331--20342, 2020.

\bibitem[Long et~al.(2018)Long, Bindschaedler, Wang, Bu, Wang, Tang, Gunter,
  and Chen]{long2018understanding}
Yunhui Long, Vincent Bindschaedler, Lei Wang, Diyue Bu, Xiaofeng Wang, Haixu
  Tang, Carl~A. Gunter, and Kai Chen.
\newblock Understanding membership inferences on well-generalized learning
  models.
\newblock \emph{arXiv preprint arXiv:1802.04889}, 2018.

\bibitem[Lukas et~al.(2023)Lukas, Salem, Sim, Tople, Wutschitz, and
  Zanella-B{\'e}guelin]{lukas2023analyzing}
Nils Lukas, Ahmed Salem, Robert Sim, Shruti Tople, Lukas Wutschitz, and
  Santiago Zanella-B{\'e}guelin.
\newblock Analyzing leakage of personally identifiable information in language
  models.
\newblock In \emph{IEEE Symposium on Security and Privacy (S\&P)}, pages
  346--363, 2023.

\bibitem[Lundberg and Lee(2017)]{lundberg2017unified}
Scott~M Lundberg and Su-In Lee.
\newblock A unified approach to interpreting model predictions.
\newblock \emph{Adv. Neural Inf. Process. Syst. (NeurIPS)}, 2017.

\bibitem[Luo and Li(2024)]{luo2024demem}
Xiaoyu Luo and Qiongxiu Li.
\newblock Demem: Privacy-enhanced robust adversarial learning via
  de-memorization.
\newblock \emph{arXiv preprint arXiv:2412.05767}, 2024.

\bibitem[Ma et~al.()Ma, Bassily, and Belkin]{ma2018power}
Siyuan Ma, Raef Bassily, and Mikhail Belkin.
\newblock The power of interpolation: Understanding the effectiveness of sgd in
  modern over-parametrized learning.
\newblock In \emph{Int. Conf. Mach. Learn. (ICML)}, pages 3325--3334.

\bibitem[Ma et~al.(2022)Ma, Wang, and Liu]{ma2022tradeoff}
Xinsong Ma, Zekai Wang, and Weiwei Liu.
\newblock On the tradeoff between robustness and fairness.
\newblock \emph{Adv. Neural Inf. Process. Syst. (NeurIPS)}, 35:\penalty0
  26230--26241, 2022.

\bibitem[Ma et~al.(2019)Ma, Zhu, and Hsu]{ma2019data}
Yuzhe Ma, Xiaojin Zhu, and Justin Hsu.
\newblock Data poisoning against differentially-private learners: Attacks and
  defenses.
\newblock \emph{arXiv preprint arXiv:1903.09860}, 2019.

\bibitem[M{\k{a}}dry et~al.(2017)M{\k{a}}dry, Makelov, Schmidt, Tsipras, and
  Vladu]{mkadry2017towards}
Aleksander M{\k{a}}dry, Aleksandar Makelov, Ludwig Schmidt, Dimitris Tsipras,
  and Adrian Vladu.
\newblock Towards deep learning models resistant to adversarial attacks.
\newblock \emph{stat}, 1050\penalty0 (9), 2017.

\bibitem[Maini et~al.(2023)Maini, Mozer, Sedghi, Lipton, Kolter, and
  Zhang]{maini2023can}
Pratyush Maini, Michael~C Mozer, Hanie Sedghi, Zachary~C Lipton, J~Zico Kolter,
  and Chiyuan Zhang.
\newblock Can neural network memorization be localized?
\newblock \emph{arXiv preprint arXiv:2307.09542}, 2023.

\bibitem[Mangalam and Prabhu(2019)]{mangalam2019deep}
Karttikeya Mangalam and Vinay~Uday Prabhu.
\newblock Do deep neural networks learn shallow learnable examples first?
\newblock In \emph{Int. Conf. Mach. Learn. Worksh.(ICML Workshop)}, 2019.

\bibitem[Mantelero(2013)]{mantelero2013eu}
Alessandro Mantelero.
\newblock The eu proposal for a general data protection regulation and the
  roots of the ‘right to be forgotten’.
\newblock \emph{Computer Law \& Security Review}, 29\penalty0 (3):\penalty0
  229--235, 2013.

\bibitem[Matsumoto et~al.(2023)Matsumoto, Miura, and
  Yanai]{matsumoto2023membership}
Tomoya Matsumoto, Takayuki Miura, and Naoto Yanai.
\newblock Membership inference attacks against diffusion models.
\newblock In \emph{Security and Privacy Workshops}, pages 77--83. IEEE, 2023.

\bibitem[Mattern et~al.(2023)Mattern, Mireshghallah, Jin, Schoelkopf, Sachan,
  and Berg-Kirkpatrick]{mattern2023membership}
Justus Mattern, Fatemehsadat Mireshghallah, Zhijing Jin, Bernhard Schoelkopf,
  Mrinmaya Sachan, and Taylor Berg-Kirkpatrick.
\newblock Membership inference attacks against language models via
  neighbourhood comparison.
\newblock In \emph{Findings of the Association for Computational Linguistics:
  ACL 2023}, pages 11330--11343, 2023.

\bibitem[McAllester(1998)]{mcallester1998some}
David~A McAllester.
\newblock Some pac-bayesian theorems.
\newblock In \emph{Proceedings of the eleventh annual conference on
  Computational learning theory}, pages 230--234, 1998.

\bibitem[Mehrabi et~al.(2021)Mehrabi, Morstatter, Saxena, Lerman, and
  Galstyan]{mehrabi2021survey}
Ninareh Mehrabi, Fred Morstatter, Nripsuta Saxena, Kristina Lerman, and Aram
  Galstyan.
\newblock A survey on bias and fairness in machine learning.
\newblock volume~54. ACM, 2021.

\bibitem[Mehta et~al.(2021)Mehta, Cutkosky, and Neyshabur]{mehtaextreme}
Harsh Mehta, Ashok Cutkosky, and Behnam Neyshabur.
\newblock Extreme memorization via scale of initialization.
\newblock In \emph{Int. Conf. Learn. Represent. (ICLR)}, 2021.

\bibitem[Mejia et~al.(2019)Mejia, Gamble, Hampel-Arias, Lomnitz, Lopatina,
  Tindall, and Barrios]{mejia2019robust}
Felipe~A Mejia, Paul Gamble, Zigfried Hampel-Arias, Michael Lomnitz, Nina
  Lopatina, Lucas Tindall, and Maria~Alejandra Barrios.
\newblock Robust or private? adversarial training makes models more vulnerable
  to privacy attacks.
\newblock \emph{arXiv preprint arXiv:1906.06449}, 2019.

\bibitem[Mohri(2018)]{mohri2018foundations}
Mehryar Mohri.
\newblock Foundations of machine learning, 2018.

\bibitem[Nagarajan et~al.(2018)Nagarajan, Raffel, and
  Goodfellow]{nagarajan2018theoretical}
Vaishnavh Nagarajan, Colin Raffel, and Ian~J Goodfellow.
\newblock Theoretical insights into memorization in gans.
\newblock In \emph{NeurIPS Workshop}, volume~1, page~3, 2018.

\bibitem[Nanda et~al.(2021)Nanda, Dooley, Singla, Feizi, and
  Dickerson]{nanda2021fairness}
Vedant Nanda, Samuel Dooley, Sahil Singla, Soheil Feizi, and John~P Dickerson.
\newblock Fairness through robustness: Investigating robustness disparity in
  deep learning.
\newblock In \emph{FAccT}, pages 466--477. ACM, 2021.

\bibitem[Newsome et~al.(2006)Newsome, Karp, and Song]{newsome2006paragraph}
James Newsome, Brad Karp, and Dawn Song.
\newblock Paragraph: Thwarting signature learning by training maliciously.
\newblock In \emph{Recent Advances in Intrusion Detection: 9th International
  Symposium, RAID 2006 Hamburg, Germany, September 20-22, 2006 Proceedings 9},
  pages 81--105. Springer, 2006.

\bibitem[Nie et~al.(2022)Nie, Guo, Huang, Xiao, Vahdat, and
  Anandkumar]{Nie2022diffpure}
Weill Nie, Brandon Guo, Yujia Huang, Chaowei Xiao, Arash Vahdat, and Animashree
  Anandkumar.
\newblock Diffusion models for adversarial purification.
\newblock In \emph{Int. Conf. Mach. Learn. (ICML)}, volume 162, pages
  16805--16827, 2022.

\bibitem[Nigatu et~al.(2024)Nigatu, Tonja, Rosman, Solorio, and
  Choudhury]{nigatu-etal-2024-zenos}
Hellina~Hailu Nigatu, Atnafu~Lambebo Tonja, Benjamin Rosman, Thamar Solorio,
  and Monojit Choudhury.
\newblock The zeno`s paradox of {\textquoteleft}low-resource' languages.
\newblock In Yaser Al-Onaizan, Mohit Bansal, and Yun-Nung Chen, editors,
  \emph{Proceedings of the 2024 Conference on Empirical Methods in Natural
  Language Processing}, pages 17753--17774, 2024.

\bibitem[Oneto et~al.(2022)Oneto, Navarin, Biggio, Errica, Micheli, Scarselli,
  Bianchini, Demetrio, Bongini, Tacchella, et~al.]{oneto2022towards}
Luca Oneto, Nicolo Navarin, Battista Biggio, Federico Errica, Alessio Micheli,
  Franco Scarselli, Monica Bianchini, Luca Demetrio, Pietro Bongini, Armando
  Tacchella, et~al.
\newblock Towards learning trustworthily, automatically, and with guarantees on
  graphs: An overview.
\newblock \emph{Neurocomputing}, 493:\penalty0 217--243, 2022.

\bibitem[{P. Paillier}(1999)]{paillier1999public}
{P. Paillier}.
\newblock Public-key cryptosystems based on composite degree residuosity
  classes.
\newblock In \emph{EUROCRYPT, pp. 223--238}, 1999.

\bibitem[Pannekoek and Spigler(2021)]{pannekoek2021investigating}
Marlotte Pannekoek and Giacomo Spigler.
\newblock Investigating trade-offs in utility, fairness and differential
  privacy in neural networks.
\newblock \emph{arXiv preprint arXiv:2102.05975}, 2021.

\bibitem[Papernot et~al.(2016)Papernot, Abadi, Erlingsson, Goodfellow, and
  Talwar]{papernot2016semi}
Nicolas Papernot, Mart{\'\i}n Abadi, Ulfar Erlingsson, Ian Goodfellow, and
  Kunal Talwar.
\newblock Semi-supervised knowledge transfer for deep learning from private
  training data.
\newblock \emph{arXiv preprint arXiv:1610.05755}, 2016.

\bibitem[Papernot et~al.(2018)Papernot, McDaniel, Sinha, and
  Wellman]{papernot2018sok}
Nicolas Papernot, Patrick McDaniel, Arunesh Sinha, and Michael~P Wellman.
\newblock Sok: Security and privacy in machine learning.
\newblock In \emph{EuroS\&P}, pages 399--414. IEEE, 2018.

\bibitem[Pardau(2018)]{pardau2018california}
Stuart~L Pardau.
\newblock The california consumer privacy act: Towards a european-style privacy
  regime in the united states.
\newblock \emph{J. Tech. L. \& Pol'y}, 23:\penalty0 68, 2018.

\bibitem[Pfeiffer et~al.(2022)Pfeiffer, Goyal, Lin, Li, Cross, Riedel, and
  Artetxe]{pfeiffer-etal-2022-lifting}
Jonas Pfeiffer, Naman Goyal, Xi~Lin, Xian Li, James Cross, Sebastian Riedel,
  and Mikel Artetxe.
\newblock Lifting the curse of multilinguality by pre-training modular
  transformers.
\newblock In Marine Carpuat, Marie-Catherine de~Marneffe, and Ivan~Vladimir
  Meza~Ruiz, editors, \emph{Proceedings of the 2022 Conference of the North
  American Chapter of the Association for Computational Linguistics: Human
  Language Technologies}, pages 3479--3495, 2022.

\bibitem[Pikuliak et~al.(2021)Pikuliak, {\v{S}}imko, and
  Bielikov{\'a}]{pikuliak2021cross}
Mat{\'u}{\v{s}} Pikuliak, Mari{\'a}n {\v{S}}imko, and M{\'a}ria Bielikov{\'a}.
\newblock Cross-lingual learning for text processing: A survey.
\newblock \emph{Expert Systems with Applications}, 165:\penalty0 113765, 2021.

\bibitem[Qiu et~al.(2023)Qiu, Ziser, Korhonen, Ponti, and
  Cohen]{qiu2023detecting}
Yifu Qiu, Yftah Ziser, Anna Korhonen, Edoardo~M Ponti, and Shay~B Cohen.
\newblock Detecting and mitigating hallucinations in multilingual
  summarisation.
\newblock \emph{arXiv preprint arXiv:2305.13632}, 2023.

\bibitem[R.~Shokri and Shmatikov(2017)]{shokri2017membership}
C.~Song R.~Shokri, M.~Stronati and V.~Shmatikov.
\newblock Membership inference attacks against machine learning models.
\newblock In \emph{IEEE Symposium on Security and Privacy (S\&P)}, pages 3--18,
  2017.

\bibitem[Ravikumar et~al.(2024)Ravikumar, Soufleri, Hashemi, and
  Roy]{ravikumar2024unveiling}
Deepak Ravikumar, Efstathia Soufleri, Abolfazl Hashemi, and Kaushik Roy.
\newblock Unveiling privacy, memorization, and input curvature links.
\newblock \emph{arXiv preprint arXiv:2402.18726}, 2024.

\bibitem[Rezaei and Liu(2021)]{rezaei2021difficulty}
Shahbaz Rezaei and Xin Liu.
\newblock On the difficulty of membership inference attacks.
\newblock In \emph{IEEE Conf. Comput. Vis. Pattern Recog. (CVPR)}, pages
  7892--7900, 2021.

\bibitem[Ribeiro et~al.(2016)Ribeiro, Singh, and Guestrin]{ribeiro2016should}
Marco~Tulio Ribeiro, Sameer Singh, and Carlos Guestrin.
\newblock " why should i trust you?": Explaining the predictions of any
  classifier.
\newblock In \emph{SIGKDD}. ACM, 2016.

\bibitem[Rice et~al.(2020)Rice, Wong, and Kolter]{rice2020overfitting}
Leslie Rice, Eric Wong, and Zico Kolter.
\newblock Overfitting in adversarially robust deep learning.
\newblock In \emph{Int. Conf. Mach. Learn. (ICML)}, pages 8093--8104, 2020.

\bibitem[Roh et~al.(2021)Roh, Lee, Whang, and Suh]{roh2021sample}
Yuji Roh, Kangwook Lee, Steven Whang, and Changho Suh.
\newblock Sample selection for fair and robust training.
\newblock \emph{Adv. Neural Inf. Process. Syst. (NeurIPS)}, 34:\penalty0
  815--827, 2021.

\bibitem[Ross and Doll{\'a}r(2017)]{ross2017focal}
T-YLPG Ross and GKHP Doll{\'a}r.
\newblock Focal loss for dense object detection.
\newblock In \emph{IEEE Conf. Comput. Vis. Pattern Recog. (CVPR)}, pages
  2980--2988, 2017.

\bibitem[Sablayrolles et~al.(2019)Sablayrolles, Douze, Schmid, Ollivier, and
  J{\'{e}}gou]{sablayrolles2019white}
Alexandre Sablayrolles, Matthijs Douze, Cordelia Schmid, Yann Ollivier, and
  Herv{\'{e}} J{\'{e}}gou.
\newblock White-box vs black-box: Bayes optimal strategies for membership
  inference.
\newblock In \emph{Int. Conf. Mach. Learn. (ICML)}, pages 5558--5567, 2019.

\bibitem[Salem et~al.(2019)Salem, Zhang, Humbert, Berrang, Fritz, and
  Backes]{salem2018ml}
Ahmed Salem, Yang Zhang, Mathias Humbert, Pascal Berrang, Mario Fritz, and
  Michael Backes.
\newblock Ml-leaks: Model and data independent membership inference attacks and
  defenses on machine learning models.
\newblock \emph{Network and Distributed Systems Security Symposium}, 2019.

\bibitem[Sankar et~al.(2025)Sankar, Kosut, Calmon, Ozgur, Wang, Shayevitz, and
  Sadeghi]{sankar2025jsait}
Lalitha Sankar, Oliver Kosut, Flavio Calmon, Ayfer Ozgur, Lele Wang, Ofer
  Shayevitz, and Parastoo Sadeghi.
\newblock Jsait issue on information-theoretic methods for trustworthy and
  reliable machine learning.
\newblock \emph{IEEE J. Sel. Areas Inf. Theory. (JSAIT)}, 5:\penalty0 xii--xv,
  2025.

\bibitem[Sanyal et~al.(2020)Sanyal, Dokania, Kanade, and
  Torr]{sanyal2020benign}
Amartya Sanyal, Puneet~K Dokania, Varun Kanade, and Philip~HS Torr.
\newblock How benign is benign overfitting?
\newblock \emph{arXiv preprint arXiv:2007.04028}, 2020.

\bibitem[Satvaty et~al.(2024)Satvaty, Verberne, and
  Turkmen]{satvaty2024undesirable}
Ali Satvaty, Suzan Verberne, and Fatih Turkmen.
\newblock Undesirable memorization in large language models: A survey.
\newblock \emph{arXiv preprint arXiv:2410.02650}, 2024.

\bibitem[Shafahi et~al.(2018)Shafahi, Huang, Najibi, Suciu, Studer, Dumitras,
  and Goldstein]{shafahi2018poison}
Ali Shafahi, W~Ronny Huang, Mahyar Najibi, Octavian Suciu, Christoph Studer,
  Tudor Dumitras, and Tom Goldstein.
\newblock Poison frogs! targeted clean-label poisoning attacks on neural
  networks.
\newblock \emph{Adv. Neural Inf. Process. Syst. (NeurIPS)}, 31, 2018.

\bibitem[Shaham et~al.(2023)Shaham, Hajisafi, Quan, Nguyen, Krishnamachari,
  Peris, Ghinita, Shahabi, and Pathirana]{shaham2023holistic}
Sina Shaham, Arash Hajisafi, Minh~K Quan, Dinh~C Nguyen, Bhaskar
  Krishnamachari, Charith Peris, Gabriel Ghinita, Cyrus Shahabi, and Pubudu~N
  Pathirana.
\newblock Holistic survey of privacy and fairness in machine learning.
\newblock \emph{arXiv preprint arXiv:2307.15838}, 2023.

\bibitem[Shen et~al.(2016)Shen, Lin, and Huang]{shen2016relay}
Li~Shen, Zhouchen Lin, and Qingming Huang.
\newblock Relay backpropagation for effective learning of deep convolutional
  neural networks.
\newblock In \emph{Computer Vision--ECCV 2016: 14th European Conference,
  Amsterdam, The Netherlands, October 11--14, 2016, Proceedings, Part VII 14},
  pages 467--482. Springer, 2016.

\bibitem[Shen et~al.(2024)Shen, Tan, Chen, Chen, Zhang, Xu, Zheng, Koehn, and
  Khashabi]{shen_language_2024}
Lingfeng Shen, Weiting Tan, Sihao Chen, Yunmo Chen, Jingyu Zhang, Haoran Xu,
  Boyuan Zheng, Philipp Koehn, and Daniel Khashabi.
\newblock The {Language} {Barrier}: {Dissecting} {Safety} {Challenges} of
  {LLMs} in {Multilingual} {Contexts}.
\newblock In \emph{Findings of the {Association} for {Computational}
  {Linguistics} {ACL} 2024}, pages 2668--2680, Bangkok, Thailand and virtual
  meeting, 2024. Association for Computational Linguistics.
\newblock \doi{10.18653/v1/2024.findings-acl.156}.
\newblock URL \url{https://aclanthology.org/2024.findings-acl.156}.

\bibitem[Shi et~al.(2021)Shi, Holtz, and Mishne]{Shi2021purification}
Changhao Shi, Chester Holtz, and Gal Mishne.
\newblock Online adversarial purification based on self-supervised learning.
\newblock In \emph{Int. Conf. Learn. Represent. (ICLR)}, 2021.

\bibitem[Simonyan et~al.(2013)Simonyan, Vedaldi, and
  Zisserman]{simonyan2013deep}
Karen Simonyan, Andrea Vedaldi, and Andrew Zisserman.
\newblock Deep inside convolutional networks: Visualising image classification
  models and saliency maps.
\newblock \emph{arXiv preprint arXiv:1312.6034}, 2013.

\bibitem[Song and Raghunathan(2020)]{song2020informationleakageembeddingmodels}
Congzheng Song and Ananth Raghunathan.
\newblock Information leakage in embedding models.
\newblock \emph{arXiv preprint arXiv:2004.00053}, 2020.

\bibitem[Song et~al.(2024)Song, Huang, Zhou, and Ma]{song_multilingual_2024}
Jiayang Song, Yuheng Huang, Zhehua Zhou, and Lei Ma.
\newblock Multilingual blending: Llm safety alignment evaluation with language
  mixture.
\newblock \emph{arXiv preprint arXiv:2407.07342}, 2024.

\bibitem[Song et~al.(2019)Song, Shokri, and Mittal]{song2019privacy}
Liwei Song, Reza Shokri, and Prateek Mittal.
\newblock Privacy risks of securing machine learning models against adversarial
  examples.
\newblock In \emph{ACM SIGSAC Conference on Computer and Communications
  Security (CCS)}, pages 241--257, 2019.

\bibitem[Stephenson et~al.(2021)Stephenson, Padhy, Ganesh, Hui, Tang, and
  Chung]{stephenson2021geometry}
Cory Stephenson, Suchismita Padhy, Abhinav Ganesh, Yue Hui, Hanlin Tang, and
  SueYeon Chung.
\newblock On the geometry of generalization and memorization in deep neural
  networks.
\newblock \emph{arXiv preprint arXiv:2105.14602}, 2021.

\bibitem[Suriyakumar et~al.(2021)Suriyakumar, Papernot, Goldenberg, and
  Ghassemi]{suriyakumar2021chasing}
Vinith~M Suriyakumar, Nicolas Papernot, Anna Goldenberg, and Marzyeh Ghassemi.
\newblock Chasing your long tails: Differentially private prediction in health
  care settings.
\newblock In \emph{FAccT}, pages 723--734. ACM, 2021.

\bibitem[{T. M. Cover and J. A. Tomas}(2012)]{cover2012elements}
{T. M. Cover and J. A. Tomas}.
\newblock \emph{Elements of information theory}.
\newblock John Wiley \& Sons, 2012.

\bibitem[Tan et~al.(2020)Tan, Wang, Li, Li, Ouyang, Yin, and
  Yan]{tan2020equalization}
Jingru Tan, Changbao Wang, Buyu Li, Quanquan Li, Wanli Ouyang, Changqing Yin,
  and Junjie Yan.
\newblock Equalization loss for long-tailed object recognition.
\newblock In \emph{IEEE Conf. Comput. Vis. Pattern Recog. (CVPR)}, pages
  11662--11671, 2020.

\bibitem[Tao et~al.(2021)Tao, Feng, Yi, Huang, and Chen]{tao2021better}
Lue Tao, Lei Feng, Jinfeng Yi, Sheng-Jun Huang, and Songcan Chen.
\newblock Better safe than sorry: Preventing delusive adversaries with
  adversarial training.
\newblock \emph{Adv. Neural Inf. Process. Syst. (NeurIPS)}, 34:\penalty0
  16209--16225, 2021.

\bibitem[Thakur et~al.(2024)Thakur, Bonifacio, Zhang, Ogundepo, Kamalloo,
  Hermelo, Li, Liu, Chen, Rezagholizadeh, et~al.]{thakur2024knowing}
Nandan Thakur, Luiz Bonifacio, Crystina Zhang, Odunayo Ogundepo, Ehsan
  Kamalloo, David~Alfonso Hermelo, Xiaoguang Li, Qun Liu, Boxing Chen, Mehdi
  Rezagholizadeh, et~al.
\newblock “knowing when you don’t know”: A multilingual relevance
  assessment dataset for robust retrieval-augmented generation.
\newblock In \emph{Findings of the Association for Computational Linguistics:
  EMNLP}, pages 12508--12526, 2024.

\bibitem[Tian et~al.(2024)Tian, Zhang, Liu, Zhu, Ding, and
  Zhou]{tian2024fairness}
Huan Tian, Guangsheng Zhang, Bo~Liu, Tianqing Zhu, Ming Ding, and Wanlei Zhou.
\newblock When fairness meets privacy: exploring privacy threats in fair binary
  classifiers via membership inference attacks.
\newblock In \emph{Int. Joint Conf. Artif. Intell. (IJCAI)}, 2024.

\bibitem[Toneva et~al.(2019)Toneva, Sordoni, des Combes, Trischler, Bengio, and
  Gordon]{toneva2019empirical}
Mariya Toneva, Alessandro Sordoni, Remi~Tachet des Combes, Adam Trischler,
  Yoshua Bengio, and Geoff Gordon.
\newblock An empirical study of example forgetting during deep neural network
  learning.
\newblock \emph{Int. Conf. Learn. Represent. (ICLR)}, 2019.

\bibitem[Touvron et~al.(2023)Touvron, Lavril, Izacard, Martinet, Lachaux,
  Lacroix, Rozière, Goyal, Hambro, Azhar, Rodriguez, Joulin, Grave, and
  Lample]{touvron2023llamaopenefficientfoundation}
Hugo Touvron, Thibaut Lavril, Gautier Izacard, Xavier Martinet, Marie-Anne
  Lachaux, Timothée Lacroix, Baptiste Rozière, Naman Goyal, Eric Hambro,
  Faisal Azhar, Aurelien Rodriguez, Armand Joulin, Edouard Grave, and Guillaume
  Lample.
\newblock Llama: Open and efficient foundation language models.
\newblock \emph{arXiv preprint arXiv:2302.13971}, 2023.

\bibitem[Tram{\`e}r et~al.(2022)Tram{\`e}r, Shokri, San~Joaquin, Le, Jagielski,
  Hong, and Carlini]{tramer2022truth}
Florian Tram{\`e}r, Reza Shokri, Ayrton San~Joaquin, Hoang Le, Matthew
  Jagielski, Sanghyun Hong, and Nicholas Carlini.
\newblock Truth serum: Poisoning machine learning models to reveal their
  secrets.
\newblock In \emph{ACM SIGSAC Conference on Computer and Communications
  Security (CCS)}, pages 2779--2792, 2022.

\bibitem[Tran et~al.(2021)Tran, Dinh, and Fioretto]{tran2021differentially}
Cuong Tran, My~Dinh, and Ferdinando Fioretto.
\newblock Differentially private empirical risk minimization under the fairness
  lens.
\newblock \emph{Adv. Neural Inf. Process. Syst. (NeurIPS)}, 34:\penalty0
  27555--27565, 2021.

\bibitem[Tran et~al.(2023)Tran, Fioretto, Khalil, Thai, and
  Phan]{tran2023fairdp}
Khang Tran, Ferdinando Fioretto, Issa Khalil, My~T Thai, and NhatHai Phan.
\newblock Fairdp: Certified fairness with differential privacy.
\newblock \emph{arXiv preprint arXiv:2305.16474}, 2023.

\bibitem[Tsipras et~al.(2018)Tsipras, Santurkar, Engstrom, Turner, and
  Madry]{tsipras2018robustness}
Dimitris Tsipras, Shibani Santurkar, Logan Engstrom, Alexander Turner, and
  Aleksander Madry.
\newblock Robustness may be at odds with accuracy.
\newblock \emph{arXiv preprint arXiv:1805.12152}, 2018.

\bibitem[Tursynbek et~al.(2020)Tursynbek, Petiushko, and
  Oseledets]{tursynbek2020robustness}
Nurislam Tursynbek, Aleksandr Petiushko, and Ivan Oseledets.
\newblock Robustness threats of differential privacy.
\newblock \emph{arXiv preprint arXiv:2012.07828}, 2020.

\bibitem[Uniyal et~al.(2021)Uniyal, Naidu, Kotti, Singh, Kenfack,
  Mireshghallah, and Trask]{uniyal2021dp}
Archit Uniyal, Rakshit Naidu, Sasikanth Kotti, Sahib Singh, Patrik~Joslin
  Kenfack, Fatemehsadat Mireshghallah, and Andrew Trask.
\newblock Dp-sgd vs pate: Which has less disparate impact on model accuracy?
\newblock \emph{arXiv preprint arXiv:2106.12576}, 2021.

\bibitem[Upadhayay and Behzadan(2024)]{upadhayay_sandwich_2024}
Bibek Upadhayay and Vahid Behzadan.
\newblock Sandwich attack: {Multi}-language {Mixture} {Adaptive} {Attack} on
  {LLMs}.
\newblock In \emph{Proceedings of the 4th {Workshop} on {Trustworthy} {Natural}
  {Language} {Processing} ({TrustNLP} 2024)}, pages 208--226, Mexico City,
  Mexico, 2024. Association for Computational Linguistics.
\newblock \doi{10.18653/v1/2024.trustnlp-1.18}.
\newblock URL \url{https://aclanthology.org/2024.trustnlp-1.18}.

\bibitem[Upreti et~al.(2024)Upreti, Lind, Elmokashfi, and
  Yazidi]{upreti2024trustworthy}
Ramesh Upreti, Pedro~G Lind, Ahmed Elmokashfi, and Anis Yazidi.
\newblock Trustworthy machine learning in the context of security and privacy.
\newblock \emph{International Journal of Information Security}, 23\penalty0
  (3):\penalty0 2287--2314, 2024.

\bibitem[Van et~al.(2022)Van, Du, Wu, and Lu]{van2022poisoning}
Minh-Hao Van, Wei Du, Xintao Wu, and Aidong Lu.
\newblock Poisoning attacks on fair machine learning.
\newblock In \emph{International Conference on Database Systems for Advanced
  Applications}, pages 370--386. Springer, 2022.

\bibitem[Van~Horn and Perona(2017)]{van2017devil}
Grant Van~Horn and Pietro Perona.
\newblock The devil is in the tails: Fine-grained classification in the wild.
\newblock \emph{arXiv preprint arXiv:1709.01450}, 2017.

\bibitem[Vapnik(2013)]{vapnik2013nature}
Vladimir Vapnik.
\newblock \emph{The nature of statistical learning theory}.
\newblock Springer science \& business media, 2013.

\bibitem[Vapnik et~al.(1994)Vapnik, Levin, and Le~Cun]{vapnik1994measuring}
Vladimir Vapnik, Esther Levin, and Yann Le~Cun.
\newblock Measuring the vc-dimension of a learning machine.
\newblock \emph{Neural computation}, 6\penalty0 (5):\penalty0 851--876, 1994.

\bibitem[Walter and Suina(2023)]{walter2023indigenous}
Maggie Walter and Michele Suina.
\newblock Indigenous data, indigenous methodologies and indigenous data
  sovereignty.
\newblock In \emph{Educational Research Practice in Southern Contexts}, pages
  207--220. Routledge, 2023.

\bibitem[Wang et~al.(2024)Wang, Xu, He, Rubinstein, and
  Cohn]{wang-etal-2024-backdoor}
Jun Wang, Qiongkai Xu, Xuanli He, Benjamin Rubinstein, and Trevor Cohn.
\newblock Backdoor attacks on multilingual machine translation.
\newblock In Kevin Duh, Helena Gomez, and Steven Bethard, editors,
  \emph{Proceedings of the 2024 Conference of the North American Chapter of the
  Association for Computational Linguistics: Human Language Technologies
  (Volume 1: Long Papers)}, pages 4515--4534, 2024.

\bibitem[Wang et~al.(2022)Wang, Xu, Liu, Li, Thuraisingham, and
  Tang]{wang2022imbalanced}
Wentao Wang, Han Xu, Xiaorui Liu, Yaxin Li, Bhavani Thuraisingham, and Jiliang
  Tang.
\newblock Imbalanced adversarial training with reweighting.
\newblock In \emph{IEEE Int. Conf. Data Mining (ICDM)}, pages 1209--1214. IEEE,
  2022.

\bibitem[Wang et~al.(2019)Wang, Song, Zhang, Song, Wang, and
  Qi]{wang2019beyond}
Zhibo Wang, Mengkai Song, Zhifei Zhang, Yang Song, Qian Wang, and Hairong Qi.
\newblock Beyond inferring class representatives: User-level privacy leakage
  from federated learning.
\newblock In \emph{IEEE INFOCOM 2019-IEEE conference on computer
  communications}, pages 2512--2520, 2019.

\bibitem[Watson et~al.(2021)Watson, Guo, Cormode, and
  Sablayrolles]{watson2021importance}
Lauren Watson, Chuan Guo, Graham Cormode, and Alex Sablayrolles.
\newblock On the importance of difficulty calibration in membership inference
  attacks.
\newblock \emph{arXiv preprint arXiv:2111.08440}, 2021.

\bibitem[Wei and Liu(2024)]{wei2024trustworthy}
Wenqi Wei and Ling Liu.
\newblock Trustworthy distributed ai systems: Robustness, privacy, and
  governance.
\newblock \emph{Computing Surveys}, 2024.

\bibitem[Wu et~al.(2020)Wu, Xia, and Wang]{wu2020adversarial}
Dongxian Wu, Shu-Tao Xia, and Yisen Wang.
\newblock Adversarial weight perturbation helps robust generalization.
\newblock \emph{Adv. Neural Inf. Process. Syst. (NeurIPS)}, 33:\penalty0
  2958--2969, 2020.

\bibitem[Wu et~al.(2021)Wu, Liu, Huang, Wang, and Lin]{wu2021adversarial}
Tong Wu, Ziwei Liu, Qingqiu Huang, Yu~Wang, and Dahua Lin.
\newblock Adversarial robustness under long-tailed distribution.
\newblock In \emph{IEEE Conf. Comput. Vis. Pattern Recog. (CVPR)}, pages
  8659--8668, 2021.

\bibitem[Wu et~al.(2024)Wu, Chen, Pan, Liu, Liu, Dai, Gao, Ma, Wu, Wang,
  et~al.]{wu2024deepseek}
Zhiyu Wu, Xiaokang Chen, Zizheng Pan, Xingchao Liu, Wen Liu, Damai Dai, Huazuo
  Gao, Yiyang Ma, Chengyue Wu, Bingxuan Wang, et~al.
\newblock Deepseek-vl2: Mixture-of-experts vision-language models for advanced
  multimodal understanding.
\newblock \emph{arXiv preprint arXiv:2412.10302}, 2024.

\bibitem[Xu et~al.(2021{\natexlab{a}})Xu, Du, and Wu]{xu2021removing}
Depeng Xu, Wei Du, and Xintao Wu.
\newblock Removing disparate impact on model accuracy in differentially private
  stochastic gradient descent.
\newblock In \emph{SIGKDD}, pages 1924--1932. ACM, 2021{\natexlab{a}}.

\bibitem[Xu et~al.(2021{\natexlab{b}})Xu, Liu, Li, Jain, and
  Tang]{xu2021robust}
Han Xu, Xiaorui Liu, Yaxin Li, Anil Jain, and Jiliang Tang.
\newblock To be robust or to be fair: Towards fairness in adversarial training.
\newblock In \emph{Int. Conf. Mach. Learn. (ICML)}, pages 11492--11501,
  2021{\natexlab{b}}.

\bibitem[Xu et~al.(2021{\natexlab{c}})Xu, Liu, Wang, Ding, Wu, Liu, Jain, and
  Tang]{xu2021towards}
Han Xu, Xiaorui Liu, Wentao Wang, Wenbiao Ding, Zhongqin Wu, Zitao Liu, Anil
  Jain, and Jiliang Tang.
\newblock Towards the memorization effect of neural networks in adversarial
  training.
\newblock \emph{arXiv preprint arXiv:2106.04794}, 2021{\natexlab{c}}.

\bibitem[Xu et~al.(2024)Xu, Wang, Zhou, Li, Xiao, and
  Chen]{xu-etal-2024-cognitive}
Nan Xu, Fei Wang, Ben Zhou, Bangzheng Li, Chaowei Xiao, and Muhao Chen.
\newblock Cognitive overload: Jailbreaking large language models with
  overloaded logical thinking.
\newblock In Kevin Duh, Helena Gomez, and Steven Bethard, editors,
  \emph{Findings of the Association for Computational Linguistics: NAACL 2024},
  pages 3526--3548, 2024.

\bibitem[Xu et~al.(2021{\natexlab{d}})Xu, Baracaldo, and Joshi]{xu2021privacy}
Runhua Xu, Nathalie Baracaldo, and James Joshi.
\newblock Privacy-preserving machine learning: Methods, challenges and
  directions.
\newblock \emph{arXiv preprint arXiv:2108.04417}, 2021{\natexlab{d}}.

\bibitem[Yang et~al.(2019)Yang, Zhang, Chang, and Liang]{yang2019neural}
Z.~Yang, J.~Zhang, E.~Chang, and Z.~Liang.
\newblock Neural network inversion in adversarial setting via background
  knowledge alignment.
\newblock In \emph{2019 ACM SIGSAC Conf. Comput. Commun. Secur.}, pages
  225--240, 2019.

\bibitem[Yeom et~al.(2018)Yeom, Giacomelli, Fredrikson, and
  Jha]{yeom2018privacy}
Samuel Yeom, Irene Giacomelli, Matt Fredrikson, and Somesh Jha.
\newblock Privacy risk in machine learning: Analyzing the connection to
  overfitting.
\newblock In \emph{IEEE Comput. Secur. Found. Symp. (CSF)}, pages 268--282,
  2018.

\bibitem[Yi and Wu(2019)]{yi2019probabilistic}
Kun Yi and Jianxin Wu.
\newblock Probabilistic end-to-end noise correction for learning with noisy
  labels.
\newblock In \emph{IEEE Conf. Comput. Vis. Pattern Recog. (CVPR)}, pages
  7017--7025, 2019.

\bibitem[Yong et~al.(2023)Yong, Menghini, and Bach]{yong2023low}
Zheng-Xin Yong, Cristina Menghini, and Stephen~H Bach.
\newblock Low-resource languages jailbreak gpt-4.
\newblock \emph{arXiv preprint arXiv:2310.02446}, 2023.

\bibitem[You et~al.(2025)You, Dai, Min, Sekhon, Joshi, and
  Duncan]{you2025silent}
Chenyu You, Haocheng Dai, Yifei Min, Jasjeet~S Sekhon, Sarang Joshi, and
  James~S Duncan.
\newblock The silent majority: Demystifying memorization effect in the presence
  of spurious correlations.
\newblock \emph{arXiv preprint arXiv:2501.00961}, 2025.

\bibitem[Yu et~al.(2018)Yu, Liu, Gong, and Tao]{yu2018learning}
Xiyu Yu, Tongliang Liu, Mingming Gong, and Dacheng Tao.
\newblock Learning with biased complementary labels.
\newblock In \emph{Eur. Conf. Comput. Vis. (ECCV)}, pages 68--83, 2018.

\bibitem[Yue et~al.(2024)Yue, Mou, Wang, and Zhao]{yue2024revisiting}
Xinli Yue, Ningping Mou, Qian Wang, and Lingchen Zhao.
\newblock Revisiting adversarial training under long-tailed distributions.
\newblock In \emph{IEEE Conf. Comput. Vis. Pattern Recog. (CVPR)}, pages
  24492--24501, 2024.

\bibitem[Zeng et~al.(2023)Zeng, Yue, Shang, Zhang, and
  Wang]{zeng2023adversarial}
Huimin Zeng, Zhenrui Yue, Lanyu Shang, Yang Zhang, and Dong Wang.
\newblock On adversarial robustness of demographic fairness in face attribute
  recognition.
\newblock In \emph{Int. Joint Conf. Artif. Intell. (IJCAI)}, pages 527--535,
  2023.

\bibitem[Zhang et~al.(2017)Zhang, Bengio, Hardt, Recht, and
  Vinyals]{zhang2017rethinking}
Chiyuan Zhang, Samy Bengio, Moritz Hardt, Benjamin Recht, and Oriol Vinyals.
\newblock Understanding deep learning requires rethinking generalization.
\newblock In \emph{Int. Conf. Learn. Represent. (ICLR)}, 2017.

\bibitem[Zhang et~al.(2021)Zhang, Bengio, Hardt, Recht, and
  Vinyals]{zhang2021understanding}
Chiyuan Zhang, Samy Bengio, Moritz Hardt, Benjamin Recht, and Oriol Vinyals.
\newblock Understanding deep learning (still) requires rethinking
  generalization.
\newblock \emph{Communications of the ACM}, 64\penalty0 (3):\penalty0 107--115,
  2021.

\bibitem[Zhang et~al.(2019)Zhang, Yu, Jiao, Xing, El~Ghaoui, and
  Jordan]{zhang2019theoretically}
Hongyang Zhang, Yaodong Yu, Jiantao Jiao, Eric Xing, Laurent El~Ghaoui, and
  Michael Jordan.
\newblock Theoretically principled trade-off between robustness and accuracy.
\newblock In \emph{Int. Conf. Mach. Learn. (ICML)}, pages 7472--7482, 2019.

\bibitem[Zhang et~al.(2024)Zhang, Zeng, Luo, Fu, Chen, Xu, and
  King]{zhang2024survey}
Yifei Zhang, Dun Zeng, Jinglong Luo, Xinyu Fu, Guanzhong Chen, Zenglin Xu, and
  Irwin King.
\newblock A survey of trustworthy federated learning: Issues, solutions, and
  challenges.
\newblock \emph{ACM Trans. Intell. Syst. Technol. (TIST}, 15\penalty0
  (6):\penalty0 1--47, 2024.

\bibitem[Zhang et~al.(2023)Zhang, Li, Cui, Cai, Liu, Fu, Huang, Zhao, Zhang,
  Chen, et~al.]{zhang2023siren}
Yue Zhang, Yafu Li, Leyang Cui, Deng Cai, Lemao Liu, Tingchen Fu, Xinting
  Huang, Enbo Zhao, Yu~Zhang, Yulong Chen, et~al.
\newblock Siren's song in the ai ocean: a survey on hallucination in large
  language models.
\newblock \emph{arXiv preprint arXiv:2309.01219}, 2023.

\bibitem[Zhang and Sabuncu(2018)]{zhang2018generalized}
Zhilu Zhang and Mert Sabuncu.
\newblock Generalized cross entropy loss for training deep neural networks with
  noisy labels.
\newblock \emph{Adv. Neural Inf. Process. Syst. (NeurIPS)}, 31, 2018.

\bibitem[Zhao et~al.(2019)Zhao, Asghar, Bhaskar, and Kaafar]{zhao2019inferring}
Benjamin Zi~Hao Zhao, Hassan~Jameel Asghar, Raghav Bhaskar, and Mohamed~Ali
  Kaafar.
\newblock On inferring training data attributes in machine learning models.
\newblock \emph{arXiv preprint arXiv:1908.10558}, 2019.

\bibitem[Zhao and Gordon(2022)]{zhao2022inherent}
Han Zhao and Geoffrey~J Gordon.
\newblock Inherent tradeoffs in learning fair representations.
\newblock \emph{Journal of Machine Learning Research}, 23\penalty0
  (57):\penalty0 1--26, 2022.

\bibitem[Zhao et~al.(2024)Zhao, Kurmanji, B{\u{a}}rbulescu, Triantafillou, and
  Triantafillou]{zhao2024makes}
Kairan Zhao, Meghdad Kurmanji, George-Octavian B{\u{a}}rbulescu, Eleni
  Triantafillou, and Peter Triantafillou.
\newblock What makes unlearning hard and what to do about it.
\newblock \emph{arXiv preprint arXiv:2406.01257}, 2024.

\bibitem[Zheng and Jiang(2022)]{zheng2022empirical}
Xiaosen Zheng and Jing Jiang.
\newblock An empirical study of memorization in nlp.
\newblock \emph{arXiv preprint arXiv:2203.12171}, 2022.

\bibitem[Zhong et~al.(2022)Zhong, Sun, Xu, Gong, and
  Wang]{zhong2022understanding}
Da~Zhong, Haipei Sun, Jun Xu, Neil Gong, and Wendy~Hui Wang.
\newblock Understanding disparate effects of membership inference attacks and
  their countermeasures.
\newblock In \emph{ASIACCS}, pages 959--974. ACM, 2022.

\bibitem[Zhu et~al.(2024)Zhu, Guo, Qi, Chu, Yu, and Li]{zhu2024survey}
Ronghang Zhu, Dongliang Guo, Daiqing Qi, Zhixuan Chu, Xiang Yu, and Sheng Li.
\newblock A survey of trustworthy representation learning across domains.
\newblock \emph{TKDD}, 2024.

\bibitem[Zhu et~al.(2014)Zhu, Anguelov, and Ramanan]{zhu2014capturing}
Xiangxin Zhu, Dragomir Anguelov, and Deva Ramanan.
\newblock Capturing long-tail distributions of object subcategories.
\newblock In \emph{IEEE Conf. Comput. Vis. Pattern Recog. (CVPR)}, pages
  915--922, 2014.

\bibitem[Zielinski et~al.(2020)Zielinski, Krishnan, and
  Chatterjee]{zielinski2020weak}
Piotr Zielinski, Shankar Krishnan, and Satrajit Chatterjee.
\newblock Weak and strong gradient directions: Explaining memorization,
  generalization, and hardness of examples at scale.
\newblock \emph{arXiv preprint arXiv:2003.07422}, 2020.

\bibitem[Zou et~al.(2023)Zou, Wang, Carlini, Nasr, Kolter, and Fredrikson]{gcg}
Andy Zou, Zifan Wang, Nicholas Carlini, Milad Nasr, J.~Zico Kolter, and Matt
  Fredrikson.
\newblock Universal and transferable adversarial attacks on aligned language
  models.
\newblock \emph{arXiv preprint arXiv:2307.15043}, 2023.

\end{thebibliography}

\end{document}